\documentclass[conference]{IEEEtran}

\usepackage{graphicx} 
\usepackage{fullpage}
\usepackage{amsmath}
\usepackage{amssymb}
\usepackage{amsthm}
\usepackage{tikz}
\usepackage{todonotes}
\usepackage{booktabs}
\usepackage{url}
\usepackage{subfig}
\usepackage{caption}
\usepackage{multirow}
\usepackage{hyperref}

\usepackage{float}

\usepackage[noend]{algpseudocode}
\usepackage[ruled]{algorithm}

\newcommand{\R}{\mathbb{R}}

\newcommand{\feats}{\mathcal{X}}
\newcommand{\labels}{\mathcal{Y}}
\newcommand{\dataset}{\mathcal{D}}
\newcommand{\dtrain}{\dataset_{\textit{train}}}
\newcommand{\dtest}{\dataset_{\textit{test}}}

\newcommand{\projsign}[2]{#1_{#2}}
\newcommand{\projfv}[2]{#1^{#2}}
\newcommand{\splits}[1]{\textit{splits}(#1)}
\newcommand{\gettrain}[1]{#1.\textit{train}}
\newcommand{\getstable}[1]{#1.\textit{stable}}

\newcommand{\revise}[1]{#1}

\title{Timber! Poisoning Decision Trees}
\author{\IEEEauthorblockN{Stefano Calzavara}
\IEEEauthorblockA{\textit{Università Ca' Foscari Venezia} \\
Venice, Italy \\
stefano.calzavara@unive.it}
\and
\IEEEauthorblockN{Lorenzo Cazzaro}
\IEEEauthorblockA{\textit{Università Ca' Foscari Venezia} \\
Venice, Italy \\
lorenzo.cazzaro@unive.it}
\and
\IEEEauthorblockN{Massimo Vettori}
\IEEEauthorblockA{\textit{Università Ca' Foscari Venezia} \\
Venice, Italy \\
884477@stud.unive.it}
}

\begin{document}

\maketitle

\begin{abstract}
We present Timber, the first white-box poisoning attack targeting decision trees. Timber is based on a greedy attack strategy that leverages sub-tree retraining to efficiently estimate the damage caused by poisoning a given training instance. The attack relies on a tree annotation procedure, which enables the sorting of training instances so that they are processed in increasing order of the computational cost of sub-tree retraining. This sorting yields a variant of Timber that supports an early stopping criterion, designed to make poisoning attacks more efficient and feasible on larger datasets. We also discuss an extension of Timber to traditional random forest models, which is valuable since decision trees are typically combined into ensembles to improve their predictive power. Our experimental evaluation on public datasets demonstrates that our attacks outperform existing baselines in terms of effectiveness, efficiency, or both. Moreover, we show that two representative defenses can mitigate the effect of our attacks, but fail to effectively thwart them.
\end{abstract}


\section{Introduction}
Our daily activities are becoming increasingly more reliant on machine learning, yet the trustworthiness of machine learning has been questioned from different points of views. A prominent class of threats against machine learning is represented by \emph{poisoning attacks}~\cite{CinaGDVZMOBPR23}. Poisoning attacks break the implicit assumption that data used to train machine learning models are representative of actual test data that will be seen upon model deployment. In particular, if the attacker can compromise the integrity of training data, e.g., by crafting incorrectly labeled instances, the training algorithm may operate on low-quality data yielding models with poor accuracy. The ultimate goal of a poisoning attack is to determine an effective way to pollute training data so as to force wrong model predictions at test time.

Poisoning attacks have been extensively investigated in the research literature. Multiple papers proposed poisoning attacks against different types of machine learning models, including support vector machines~\cite{BiggioCFGR11, XiaoXE12, XiaoBNXER15, PaudiceML18}, linear classifiers~\cite{AwasthiBL17} and neural networks~\cite{ZhangBHRV17, ZhangHSW20}. Unfortunately, prior research largely neglected the investigation of poisoning attacks against \emph{decision trees}, which are still one of the most effective machine learning models operating on tabular data~\cite{GrinsztajnOV22}. Decision trees are peculiar because they are non-differentiable models, meaning that the loss function that they optimize does not have a gradient. This implies that poisoning attacks against decision trees cannot be formalized in terms of a traditional bilevel optimization problem in the style of~\cite{CinaGDVZMOBPR23} and new custom attack algorithms must be designed. A possible way around this limitation is using \emph{black-box} attack strategies, which are model-agnostic and proved effective in some practical settings~\cite{ZhangCZL21, PrudHommeK21, TaltySB21, ShahidIWIA22, abs-1712-05526, SchusterSTS21}. Unfortunately, black-box attack strategies assume that the attacker does not know anything about the training process, hence they may underestimate the attacker's capabilities. In-depth understanding of poisoning attacks against decision trees requires the careful design and evaluation of \emph{white-box} attack strategies, where the attacker abuses the inner workings of the tree learning algorithm to their advantage. This way, we may be able to perform a conservative security analysis which takes into account more powerful attackers with additional information about their target.

\paragraph*{Contributions}
Our contributions are as follows:
\begin{enumerate}
    \item We propose Timber, the first white-box poisoning attack designed to target decision trees. Timber is based on a greedy attack strategy that leverages sub-tree retraining to efficiently estimate the damage caused by poisoning a given training instance. Timber relies on a tree annotation procedure which enables sorting training instances so that they are processed in increasing order of computational cost of sub-tree retraining. This sorting yields a variant of Timber that supports an early stopping criterion designed to make poisoning attacks more efficient and feasible on larger datasets.

    \item We discuss how to generalize Timber from individual decision trees to decision tree ensembles, in particular traditional random forest models based on independently trained trees~\cite{Breiman01}. This generalization is useful because decision trees are rarely used in isolation and ensembles of decision trees are normally used to solve challenging classification problems.

    \item We experimentally assess the performance of our attacks on public datasets, showing that they outperform existing baselines in terms of effectiveness, efficiency or both. We also show that two representative defenses can mitigate the effect of our attacks, but fail to effectively thwart them.
\end{enumerate}

\paragraph*{Code availability} To support reproducible research, we make our code available on GitHub~\cite{code}.

\section{Background}
We here present the main technical ingredients required to understand the rest of the paper.

\subsection{Supervised Learning}
Let $\feats \subseteq \R^d$ be a $d$-dimensional vector space of real-valued \textit{features}. An \emph{instance} $\vec{x} \in \feats$ is a $d$-dimensional feature vector $\langle x_1, x_2, \ldots, x_d \rangle$ representing an object in the feature space $\feats$. Each instance is assigned a class label $y \in \labels$ by an unknown \emph{target} function $f: \feats \rightarrow \labels$. Supervised learning algorithms automatically learn a \emph{classifier} $g: \feats \rightarrow \labels$ from a \emph{training set} of correctly labeled instances $\dtrain = \{(\vec{x}_i,f(\vec{x}_i))\}_i$, with the goal of approximating the unknown target function $f$ as accurately as possible.

The performance of classifiers is empirically estimated on a \emph{test set} of correctly labeled instances $\dtest = \{(\vec{z}_i,f(\vec{z}_i))\}_i$, disjoint from the training set, yet drawn from the same data distribution. A traditional measure to assess the performance of classifiers is called \emph{accuracy}, defined as the percentage of instances of the test set where the classifier performs a correct prediction. For simplicity, we here focus on binary classification tasks, i.e., we let $\labels = \{+1,-1\}$. This is a convenient setting to study poisoning attacks, because it allows us to represent poisoning in terms of \emph{label flipping} attacks~\cite{CinaGDVZMOBPR23}, where the attacker replaces a correctly labelled instance $(\vec{x}_i,f(\vec{x}_i)) \in \dtrain$ with the mislabelled instance  $(\vec{x}_i,-f(\vec{x}_i))$.

\subsection{Decision Trees}
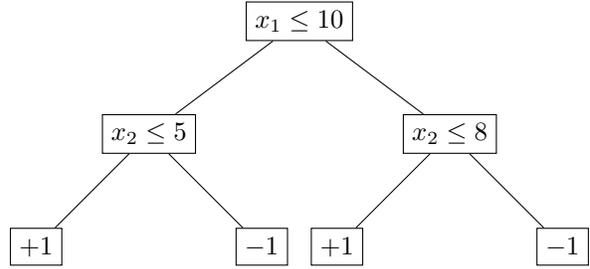
\begin{figure}[t]
\centering
\begin{tikzpicture}[level 1/.style={sibling distance=4cm},level 2/.style={sibling distance=3cm}]
\tikzstyle{every node}=[rectangle,draw]
\node{$x_1 \leq 10$}
	child { node {$x_2 \leq 5$}
	        child { node {$+1$}}
	        child { node {$-1$}} }
	child { node {$x_2 \leq 8$}
		    child { node {$+1$}}
	        child { node {$-1$}} }
;
\end{tikzpicture}
\caption{Example of decision tree.}
\label{fig:tree}
\end{figure}

We focus on traditional \emph{binary decision trees} for classification~\cite{BreimanFOS84}. Decision trees can be inductively defined as follows: a decision tree $t$ is either a leaf $\lambda(y)$ for some label $y \in \labels$ or an internal node $\sigma(f,v,t_l,t_r)$, where $f \in \{1,\ldots,d\}$ identifies a feature, $v \in \R$ is a threshold for the feature, and $t_l,t_r$ are decision trees (left and right child). At test time, the instance $\vec{x}$ traverses the tree $t$ until it reaches a leaf $\lambda(y)$, which returns the prediction $y$, denoted by $t(\vec{x}) = y$. Specifically, for each traversed tree node $\sigma(f,v,t_l,t_r)$, $\vec{x}$ falls into the left sub-tree $t_l$ if $x_f \leq v$, and into the right sub-tree $t_r$ otherwise. Figure~\ref{fig:tree} represents an example decision tree of depth 2, which assigns label $+1$ to the instance $\langle 12,7 \rangle$ and label $-1$ to the instance $\langle 8,6 \rangle$. 

Decision trees are learned by an iterative process starting from a single leaf, which is grown into a full-fledged tree with the goal of minimizing the \emph{entropy} of the leaves.\footnote{Decision trees can also be trained to minimize other measures, such as \emph{Gini impurity}. Our proposal can be readily generalized to other measures with limited effort.} For any $\dataset \subseteq \dtrain$, we define its entropy $H(\dataset)$ as follows:
\begin{align*}
H(\dataset) & = -(|\projsign{\dataset}{+1}|/|\dataset| \cdot \log_2(|\projsign{\dataset}{+1}|/|\dataset|) \\ 
& \quad\quad + |\projsign{\dataset}{-1}|/|\dataset| \cdot \log_2(|\projsign{\dataset}{-1}|/|\dataset|)),
\end{align*}
where $\projsign{\dataset}{y} = \{(\vec{x},y') \in \dataset ~|~ y' = y\}$ is the restriction of $\dataset$ to the instances with label $y$.

\begin{algorithm}[t]
\caption{Training algorithm for decision trees.}
\label{alg:tree-train}
\begin{algorithmic}[1]
\Function{Tree-Train}{$\dataset$}
\State{$\textit{best-split} \gets \bot$} 
\State{$\textit{best-gain} \gets 0$} 
\For{$(f,v) \in \splits{\dataset}$} 
    \If{$G(\dataset, f, v) > \textit{best-gain}$}
        \State{$\textit{best-split} \gets (f,v)$}
        \State{$\textit{best-gain} \gets G(\dataset, f, v)$}
    \EndIf
\EndFor
\If{$\textit{best-split} = (f^*,v^*)$} 
    \State{$t_l \gets \Call{Tree-Train}{\projfv{\dataset}{f^* \leq v^*}}$}
    \State{$t_r \gets \Call{Tree-Train}{\projfv{\dataset}{f^* > v^*}}$}
    \State{\Return{$\sigma(f^*, v^*, t_l, t_r)$}}
\Else 
    \If{$|\projsign{\dataset}{+1}| \geq |\projsign{\dataset}{-1}|$} \Return{$\lambda(+1)$}
    \Else\ \Return{$\lambda(-1)$}
    \EndIf
\EndIf
\EndFunction
\end{algorithmic}
\end{algorithm}

The training algorithm $\textsc{Tree-Train}(\dataset)$ is presented in Algorithm~\ref{alg:tree-train} and is invoked with input $\dataset = \dtrain$. The algorithm splits a leaf including the data $\dataset$ by extracting a set of candidates splits, noted $\splits{\dataset}$, which may be used to grow the tree by replacing the leaf with a new decision tree of depth one. The simplest definition of $\splits{\dataset}$ is $\splits{\dataset} = \{(f,v) ~|~ \exists (\vec{x},y) \in \dataset: x_f = v\}$, but implementations may vary and we do not make any assumption on how the candidate splits are computed. The training algorithm computes, for each $(f,v) \in \splits{\dataset}$, how the entropy would change if the leaf was grown into a tree of the form $\sigma(f,v,\lambda(y_l),\lambda(y_r))$ for some $y_l,y_r$ minimizing the prediction errors in the leaves. This is done by computing the \emph{information gain} $G(\dataset,f,v)$ resulting from partitioning $\dataset$ by using the split $(f,v)$, which is defined as follows:
\begin{align*}
G(\dataset,f,v) & = H(\dataset) - (|\projfv{\dataset}{f \leq v}|/|\dataset| \cdot H(\projfv{\dataset}{f \leq v}) \\
& \quad\quad + |\projfv{\dataset}{f > v}|/|\dataset| \cdot H(\projfv{\dataset}{f > v})), 
\end{align*}
where $\projfv{\dataset}{f \leq v} = \{(\vec{x},y) \in \dataset ~|~ x_f \leq v\}$ and $\projfv{\dataset}{f > v} = \{(\vec{x},y) \in \dataset ~|~ x_f > v\}$. Once the best split $(f^*,v^*)$ has been found, the original leaf is replaced by the decision tree $\sigma(f^*,v^*,t_l,t_r)$, where $t_l$ and $t_r$ are the decision trees recursively trained over $\projfv{\dataset}{f^* \leq v^*}$ and $\projfv{\dataset}{f^* > v^*}$ respectively. The tree construction terminates when none of the possible splits enables some information gain or some other termination criterion is met, e.g., the tree exceeds a maximum depth (for simplicity, alternative termination criteria are not shown in the pseudo-code).

The computational complexity of the tree training algorithm is $O(d \cdot n^2 \log(n))$, where $d$ is the number of features and $n$ is the size of the training set~\cite{BreimanFOS84, Quinlan93}. This complexity assumes that, for each feature $f$, the information gain is computed by ordering the training data based on the value of $f$, which simplifies the computation of the partitioning induced by each candidate split $(f,v)$. In particular, each of the $d$ features requires a sorting operation of cost $O(n \cdot \log(n))$ to find the best split. This must be repeated for each node in the decision tree, whose number is bounded above by $O(n)$.

\subsection{Tree Ensembles}
Decision trees are effective models for small datasets, but they may offer suboptimal performance on large and complicated datasets. The predictive power of tree-based classifiers can be increased by training \emph{ensembles} of multiple decision trees, using algorithms like Random Forest (RF~\cite{Breiman01}) and Gradient Boosted Decision Trees (GBDT~\cite{Friedman}). RF is based on the training of multiple independent trees, each trained on a subset of the training set and a subset of the features. The ensemble prediction is then performed by aggregating individual tree predictions, e.g., using hard majority voting. GBDT instead is a more sophisticated approach in which trees are iteratively trained, with each tree $t_i$ being trained with the goal of reducing the prediction errors made by the previously trained trees $t_1, \ldots, t_{i-1}$.
\section{Poisoning Decision Trees}
We here introduce our threat model, we explain the key challenges of our research and we propose Timber, our poisoning attack operating against decision trees. We also discuss how Timber can be extended to decision tree ensembles, in particular based on the RF algorithm.

\subsection{Threat Model}
In a poisoning attack, the attacker targets the training data or the training algorithm to compromise the performance of the classifier at test time. To define our threat model for poisoning attacks, we start from a recent survey systematizing research in the field, which defines a clear attack framework and introduces terminology~\cite{CinaGDVZMOBPR23}. 

We focus on \emph{availability} violations, i.e., the attacker's goal is to decrease the accuracy of the classifier that is trained by the learning algorithm: the more the accuracy is downgraded, the more the attack is considered effective. Moreover, we focus on \emph{white-box} attacks, i.e., the attacker has complete knowledge of the training data, the training algorithm, and the model hyperparameters. In this way, we identify insights about decision tree construction that the attacker might abuse and we estimate security under the conservative assumption that the attacker has full knowledge of the training process. Finally, we assume that the attacker alters a subset of the training data collected by the target. \revise{The attacker can only modify the training labels, thus it does not perturb the features of any training sample, which is often referred to as a \emph{label flipping} attack. The attacker can flip the labels of up to $k$ arbitrarily chosen instances of the training set, leading to a \emph{poisoned} dataset which is used to train the classifier. Label flipping is an appropriate threat model for scenarios where the labeling process is adversarial. For instance, in product rating systems, an attacker may assign low scores to targeted products in a public catalog to manipulate a recommender system. Similarly, an attacker may create a rogue mailbox to mislabel spam messages as ham with the goal of fooling a remote classifier trained over user reports.}

The objective of our research is to find an algorithm to effectively identify the $k$ instances to attack out of the $n$ training instances, with the goal of compromising the accuracy of the trained classifier.

\subsection{Baselines and Challenges}
\label{sec: challenges}
We are not aware of any poisoning attack in the literature that specifically targets decision trees. A few research papers present experiments targeting decision trees (among other models) by means of model-agnostic \emph{black-box} poisoning techniques, e.g., based on the distribution of different features of the training data~\cite{ChangI2020}. Similar approaches are useful to empirically assess the dangers posed by poisoning attacks, but they make the assumption that the attacker knows nothing about the training process and do not offer any guarantees about their practical effectiveness. This motivates the importance of white-box attacks abusing the inner workings of the tree learning algorithm to magnify the advantage of the attacker and enable a conservative security analysis. There are a few white-box poisoning attacks in the literature that work for entire classes of machine learning models, such as differentiable models~\cite{XiaoXE12, PaudiceML18, AwasthiBL17, ZhangBHRV17, ZhangHSW20}. Unfortunately, these attacks do not generalize to decision trees, because decision trees are not differentiable.

We here explain why poisoning decision trees is challenging by presenting a few baseline attack methods. The first observation we make is that finding the best $k$ instances to flip by exhaustive enumeration of the subsets of instances is impossible, because there are $n \choose k$ subsets to test. Even for a small dataset of $n = 1{,}000$ instances and a tiny $k = 10$, there are around $2.634 \times 10^{23}$ available combinations, which is intractable. A possible solution is then to use a heuristic \emph{greedy} approach. We first train a decision tree $t$ over $\dtrain$ and we then try to flip each instance of $\dtrain$ before training a new tree $t'$. After trying all the instances, we flip the one leading to the tree with the lowest accuracy and we iterate the process for $k$ rounds, leading to $n \cdot k$ trees being grown. This complexity may be acceptable for small datasets, as shown in the experimental evaluation of label flipping attacks by Paudice et al.~\cite{PaudiceML18}. Unfortunately, for a medium dataset of $n = 5{,}000$ instances and $k = 500$, the greedy attack may already construct up to 2.5M trees.

To further speed up the attack, one might revise the proposed greedy approach to include an \emph{early stopping} criterion, e.g., when the attack finds any instance leading to some accuracy loss, the attack flips its label and moves to the next round. Of course, this does not change the worst-case complexity of the algorithm, but in practice this variant of the attack is expected to be much faster. To estimate the benefits of early stopping, assume that on average just 10\% of the training set must be analyzed to identify an instance leading to some accuracy loss. For a dataset of $n = 5{,}000$ instances and $k = 500$, this variant of the attack may construct around 250k trees.


\subsection{Timber: Attack Overview}
Our attack called Timber extends a traditional greedy attack strategy (possibly with early stopping) to improve its efficiency and make it usable in practice. Greedy attack strategies require the construction of a significant number of decision trees, as discussed in the previous section. The main intuition of our attack is that, if we can make decision tree construction itself more efficient, we can make greedy attack strategies scale to larger datasets. Recall that training a decision tree has a computational complexity of $O(d \cdot n^2 \log(n))$, because we split $O(n)$ nodes by paying a cost of $O(d \cdot n \log(n))$ for each node. A key insight of our attack is that the worst-case complexity of node splitting $O(d \cdot n \log(n))$ is very pessimistic, because the training set is partitioned across nodes when the tree is grown and becomes increasingly smaller, e.g., if the best split of the root is $(f,v)$, the recursive calls of the training algorithm operate over the smaller datasets $\projfv{\dataset}{f \leq v}$ and $\projfv{\dataset}{f > v}$ respectively. For example, if $(f,v)$ evenly splits the training set, the two recursive calls operate on $n/2$ instances, meaning that splitting each of the new nodes costs significantly less than splitting the root. Hence, nodes deeper in the tree are much cheaper to split than nodes higher in the tree, with the root being the most expensive node to split. 

\begin{figure}[t]
\centering
\begin{tikzpicture}[level 1/.style={sibling distance=4cm},level 2/.style={sibling distance=3cm}]
\tikzstyle{every node}=[rectangle,draw]
\tikzstyle{edge from parent}=[draw,thick,edge from parent path={(\tikzparentnode.south) -- (\tikzchildnode.north)}]

\node{$x_1 \leq 10$}
    child { node[fill=red] {$x_2 \leq 5$}
            child { node[fill=red] {$+1$}
                     edge from parent node[midway,above left,circle,inner sep=1pt] {250}}
            child { node[fill=red] {$-1$}
                     edge from parent node[midway,above right,circle,inner sep=1pt] {150}} 
    edge from parent node[midway,above left,circle,inner sep=1pt] {400}}
    child { node {$x_2 \leq 8$}
            child { node {$+1$}
                     edge from parent node[midway,above left,circle,inner sep=1pt] {350}}
            child { node {$-1$}
                     edge from parent node[midway,above right,circle,inner sep=1pt] {250}} 
    edge from parent node[midway,above right,circle,inner sep=1pt] {600}}
;
\end{tikzpicture}
\caption{Intuition of the Timber attack. If flipping the label of the instance $(\vec{x},y)$ does not invalidate the best split of the root and $\vec{x}$ falls in its left child, only the sub-tree in red (including 400 instances) may need retraining.}
\label{fig:timber}
\end{figure}
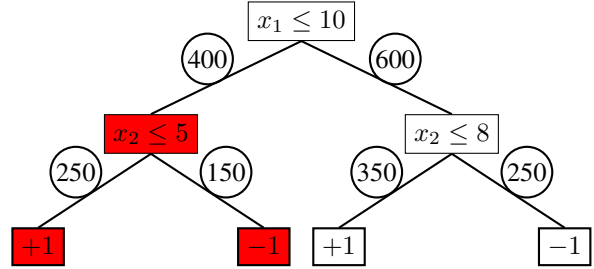

Since greedy attack strategies operate by flipping one instance at a time, the impact of this single instance on the trained tree is expected to be small in practice. In particular, flipping the label of a single training instance is unlikely to affect the best split of the root, because the best split is identified by considering $n \gg 1$ instances. However, the lower we descend in the decision tree, the higher the odds that the flipped instance affects the best split. In our example where the root evenly splits the training instances, the best split of a child of the root is found by processing just $n/2$ instances, meaning that a single label flip has a higher chance of changing the best split. This means that a single label flip normally preserves most of the structure of the decision tree and just a small, deep sub-tree where the best split has been invalidated needs to be retrained. If the root of this sub-tree includes just a small part of the training data, sub-tree retraining enables a significant speedup compared to retraining the entire tree from scratch. This intuition is shown in Figure~\ref{fig:timber}, where the numbers in the circles show how the 1,000 instances of the training set are split across the nodes upon tree construction. If flipping the label of the instance $\langle 8, 6 \rangle$ does not invalidate the best split of the root $x_1 \leq 10$, we may recursively focus on its left child (the right child can be ignored, because the poisoned instance falls on the left). Then, if the label flip invalidates the best split of the left child $x_2 \leq 5$, we only need to retrain the sub-tree in red, whose construction only involves 400 instances (40\% of the training set).

\begin{figure}[t]
    \centering
    \includegraphics[scale=0.62]{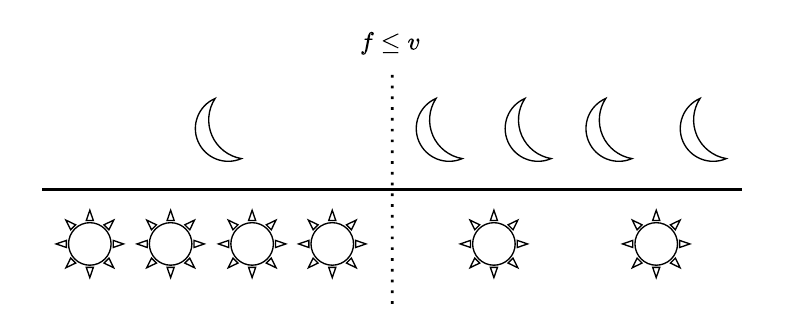}
    \caption{Splitting the dataset $\dataset$ based on the split $(f,v)$. Poisoning attacks can target positive or negative instances on the left or on the right of the split, leading to four attack possibilities that we must account for.}
    \label{fig:gains}
\end{figure}

Our attack operates by annotating each node of the decision tree with the set of the training instances which would not change the current node best split, even if their label was flipped. We refer to such instances as the \emph{stable instances} of the node. By leveraging this information, we can determine the portion of the decision tree impacted by a poisoning attack and estimate the attack effectiveness by retraining a single sub-tree, rather than the entire tree. To identify the stable instances within decision tree nodes, we leverage a compact representation of possible attacker's actions and their corresponding impact on the information gain computed during tree learning. The intuition is discussed for the dataset $\dataset$ in Figure~\ref{fig:gains}, where the sun represents instances of the positive class, the moon represents instances of the negative class, and the dotted line shows the best split $(f,v)$ of a tree node. The attacker has four possible options: $(i)$ flip a positive instance on the left of the split, $(ii)$ flip a negative instance on the left of the split, $(iii)$ flip a positive instance on the right of the split, or $(iv)$ flip a negative instance on the right of the split. In all four cases, the identity of the chosen instance is irrelevant, because the information gain depends just on the number of positive and negative instances on each side of the split. 

In our example, the initial entropy is 0.99 and the best split $(f,v)$ has an information gain of 0.16. We then observe that, if the attacker flipped a positive instance on the left of the split, the left side of the split would include three positive instances and two negative instances. In this case the entropy of the data would stay the same, but the new information gain of the split would become 0.05. We compactly represent this information with the triple $(0.05, f \leq v, +1)$, meaning that flipping a positive instance on the left of the split $(f,v)$ would lead to a new information gain of 0.05. We can similarly compute the other three triples $(0.44, f \leq v, -1), (0.31, f > v, +1), (0.07, f > v, -1)$, thus capturing the effect on the split $(f,v)$ of all the possible attacker's actions in terms of a set of four triples, denoted by $G^*(\dataset,f,v)$. Observe that $G^*(\dataset,f,v)$ includes \emph{at most} four elements, because positive or negative instances may not be present on the left or on the right of the split, meaning that some flips may be impossible. By computing $G^*(\dataset,f',v')$ for each other possible split $(f',v')$, it is possible to determine whether any attacker's action might lead to the identification of a new best split, i.e., a split with a higher information gain than $(f,v)$.

Our attack algorithm trains a decision tree over the clean training data and then operates in two steps. The first step is \emph{tree annotation} (described in Section~\ref{sec:tree-annotation}): we annotate each node of the decision tree with the set of its stable instances. The annotation process is efficient because it boils down to checking the information available in $G^*$, which can be directly computed during decision tree construction, because the training algorithm computes the information gain $G$ of all the possible splits anyways. Computing $G^*$ requires a simple adaptation of the formula used to compute $G$. The second step of the attack is \emph{label flipping} (described in Section~\ref{sec:label-flipping}): we use the computed stability information to identify the instances to prioritize in the poisoning attack to improve its efficiency. For each such instance, we flip its label and we retrain just the sub-tree of the decision tree that may be affected by this change to identify the accuracy loss. After choosing the instance to attack, e.g., the one leading to the highest accuracy loss or the first instance introducing some loss, we train a new decision tree over the poisoned dataset and we start the attack again until the maximum number of label flips has been reached.

\subsection{Tree Annotation}
\label{sec:tree-annotation}
We extend each node of the decision tree $t$ with some auxiliary information: $(i)$ the set of the training instances $\gettrain{t}$ used in the node construction, which can be readily identified by instrumenting the training algorithm, and $(ii)$ the set of the stable instances $\getstable{t}$, which is computed by the tree annotation function \textsc{Annotate} in Algorithm~\ref{alg:locally-vulnerable}. The function takes as input a decision tree and returns its annotated version. The algorithm initially assumes all the training instances to be stable and prunes the set of stable instances whenever it finds evidence that flipping a label may invalidate the best split $(f,v)$. This can only happen if there exists another split $(f',v')$ leading to a higher information gain than $(f,v)$ after label flipping, or when the information gain is the same but $(f',v')$ is processed before $(f,v)$ during tree construction. Assuming $\dataset = \gettrain{t}$, this can be determined by checking each $(g,\phi,y) \in G^*(\dataset,f,v)$ against each $(g',\phi',y') \in G^*(\dataset,f',v')$: if $g' > g$, or $g' = g$ and the split $(f',v')$ is processed before $(f,v)$ in the lexicographic order, then all the instances in the intersection $I = \projsign{\projfv{\dataset}{\phi}}{y} \cap \projsign{\projfv{\dataset}{\phi'}}{y'}$ must be removed from $V$. To understand the definition of $I$, observe that $I \neq \emptyset$ when $y = y'$ and there exist instances satisfying the predicate $\phi \wedge \phi'$, i.e., there exists a class including instances falling in the portion of the feature space common to $\phi$ and $\phi'$. For any such instance, a label flip would make $(f',v')$ the new best split in place of $(f,v)$. 

\begin{algorithm}[t]
\caption{Tree annotation algorithm}
\label{alg:locally-vulnerable}
\begin{algorithmic}[1]
\Function{Annotate}{$t$}
\State{$\getstable{t} \gets \gettrain{t}$} 
\If{$t = \sigma(f,v,t_l,t_r)$}
    \State{$\dataset \gets \gettrain{t}$} 
    \For{$(g, \phi, y) \in G^*(\dataset,f,v)$}
        \For{$(f',v') \in \splits{\dataset} \setminus \{(f,v)\}$} 
            \For{$(g', \phi', y') \in G^*(\dataset,f',v')$}
                \If{$g' > g \vee (g' = g \wedge (f',v') \prec (f,v))$} 
                    \State{$I \gets \projsign{\projfv{\dataset}{\phi}}{y} \cap \projsign{\projfv{\dataset}{\phi'}}{y'}$} 
                    \State{$\getstable{t} \gets \getstable{t} \setminus I$} 
                \EndIf
            \EndFor
        \EndFor
    \EndFor
    \State{$t_l' \gets \Call{Annotate}{t_l}$} 
    \State{$t_r' \gets \Call{Annotate}{t_r}$} 
    \State{\Return{$\sigma(f,v,t_l',t_r')$}} 
\Else
    \State{\Return{$t$}}
\EndIf
\EndFunction
\end{algorithmic}
\end{algorithm}

An important point to note is that the identification of the stable instances can be directly embedded within the tree construction at training time. Indeed, the tree construction algorithm (Algorithm~\ref{alg:tree-train}) must compute the information gain $G$ for all the possible splits anyway. We can then modify the algorithm to compute the set $G^*$ for all the possible splits, meaning that for each split we do not compute just a single information gain, but five (at most). This computation is very efficient, because it suffices to update the number of the positive and negative instances falling on the left and on the right of the split after label flipping, without any need to scan the entire dataset again. Our implementation directly integrates the computation of the stable instances in the tree construction algorithm of scikit-learn~\cite{scikit-learn}.

\subsection{Label Flipping}
\label{sec:label-flipping}
The \textsc{Flip-Retrain} function in Algorithm~\ref{alg:retrain} takes as input an already annotated decision tree $t$ and an instance $(\vec{x},y) \in \gettrain{t}$ to return the new decision tree $t'$ obtained by replacing $(\vec{x},y)$ with $(\vec{x},-y)$ in the training data. The key insight of the function is that, since we pre-computed stability information for all training instances, we can retrain just a specific sub-tree of $t$ to construct the new tree $t'$, hence $t'$ does not need to be trained from scratch. This improves efficiency because retraining operates just over a subset of the training data rather than on the entire training set. The function recursively traverses $t$ until it finds the first node where $(\vec{x},y)$ is not stable, which identifies the sub-tree of $t$ where retraining is required. Note that the retrained sub-tree must be annotated again, because its structure may have changed.

\begin{algorithm}[t]
\caption{Retraining algorithm}
\label{alg:retrain}
\begin{algorithmic}[1]
\Require{$(\vec{x},y) \in \gettrain{t}$}
\Function{Flip-Retrain}{$t, (\vec{x},y)$}
    \If{$t = \sigma(f,v,t_l,t_r) \wedge (\vec{x},y) \in \getstable{t}$}
        \If{$x_f \leq v$}
            \State{$t_l' \gets \Call{Flip-Retrain}{t_l, (\vec{x}, y)}$} 
            \State{\Return{$\sigma(f,v,t_l',t_r)$}} 
        \Else
            \State{$t_r' \gets \Call{Flip-Retrain}{t_r, (\vec{x}, y)}$}  
            \State{\Return{$\sigma(f,v,t_l,t_r')$}} 
        \EndIf
    \Else
        \State{$\dataset \gets (\gettrain{t} \setminus \{(\vec{x},y)\}) \cup \{(\vec{x},-y)\}$} 
        \State{$t \gets \Call{Tree-Train}{\dataset}$} 
        \State{\Return{\Call{Annotate}{$t$}}} 
    \EndIf
\EndFunction
\end{algorithmic}
\end{algorithm}

Of course, the use of sub-tree retraining alone does not necessarily suffice to yield an efficient poisoning attack algorithm. Indeed, although we can efficiently estimate the impact of poisoning a given instance and retrain just a sub-tree, we may still have many instances in the training set. We may then want to restrict the number of instances to consider in our poisoning attack to further speed up the process. A relevant insight here is that the stability information computed by the annotation procedure allows us to identify those instances leading to a particularly efficient sub-tree retraining, hence we may prioritize such instances in our attack strategy. The intuition is that we can assign a \emph{score} to each training instance $(\vec{x},y)$ based on the percentage of training instances included in the first node of the prediction path where $(\vec{x},y)$ is not stable. This number $s \in [0,1]$ estimates the cost of sub-tree retraining when $(\vec{x},y)$ is subject to label flipping. The score information can be used to speed up the attack by improving the efficiency of early stopping. In particular, one may sort instances based on increasing order of score, so that the attack starts from instances supporting efficient sub-tree retraining and may quickly hit the early stopping condition. 

\subsection{Extension to Tree Ensembles}
Our poisoning attack was designed and presented for traditional decision trees, however decision trees are seldom used in isolation for classification tasks due to their limited predictive power. Better classifiers can be built by training ensembles of decision trees, using algorithms like RF and GBDT. Poisoning decision tree ensembles using attack strategies like the proposed greedy approach is even more computationally expensive than targeting a single decision tree, because an ensemble may include tens or hundreds of trees to retrain, meaning that effective speedup strategies are even more important. Luckily, our proposed attack can be readily generalized to decision tree ensembles of independently trained trees like RF classifiers, because our annotation procedure can be directly applied to the individual trees constituting the ensemble. Once all the trees in the ensemble have been independently trained and annotated, we may identify the candidate instances to attack just by redefining the notion of score of an instance in terms of the mean of the scores computed for the individual trees. Intuitively, this updated notion of score estimates the aggregate cost of sub-tree retraining for all the trees in the ensemble, i.e., an instance with a small score ensures efficient sub-tree retraining in all the trees. Observe that, if the training algorithm is embarrassingly parallel like RF and there are at least as many threads as the number of trees to train, it is possible to retrain all the sub-trees in parallel, hence it might be more appropriate to replace the mean of the scores with their maximum, because the execution time of the slowest thread determines the actual execution time of the attack. The effectiveness of each label flip is estimated as the accuracy loss forced on the entire forest.

We observe that our poisoning attack cannot be readily generalized to ensembles based on interdependent trees, like GBDT models. The reason is that trees composing such models are trained sequentially, because the next tree in the ensemble is trained to minimize the prediction errors produced by the previously trained trees. Assume then that our poisoning attack is also performed sequentially and let $t_i$ be the tree under attack. Flipping the label of a training instance of $t_i$ may also affect the construction of some tree $t_j$ with $j < i$, meaning that the prediction errors performed by the previously trained trees may change, leading to the training of a different tree $t_i'$ in place of $t_i$. This means that it would be difficult to make sub-tree retraining an effective way to optimize the efficiency of the attack. We consider the generalization of our techniques to GBDT models to be an intriguing yet challenging direction for future work.
\section{Experimental Evaluation}
\label{sec:experiments}
We perform our experimental evaluation on four public datasets: Musk2~\cite{Musk2}, Wine~\cite{Wine}, Spambase~\cite{Spambase} and Breast-Cancer~\cite{Breast} (abbreviated as Breast). 
\revise{The key characteristics of the chosen datasets are reported in the appendix.} All the datasets are tabular and related to binary classification tasks, thus well-suited for decision tree learning and inference. Moreover, Musk2, Spambase and Breast-Cancer have been adopted as benchmarks in related work~\cite{PaudiceML18, BiggioNL11}. Datasets are split as 80/20 for training/testing using stratified random sampling.


    
    

\subsection{Methodology}
We assume the attacker can poison $k$ training instances, ranging from 1\% to 10\% of the training set. We consider 10\% to be an upper bound for realistic attacks. Moreover, we assume that the attacker operates over one of the two classes (the positive class). This is a realistic assumption because the attacker may be more interested in disrupting the detection of a specific class, e.g., the classification of spam emails as spam. Additionally, targeting a specific class may also harm the classification of instances of the other class.

Recall that we are not aware of any poisoning attacks designed to target decision trees, except those proposed
in this paper. Our baselines are then general attack strategies that may be applied to any type of classifier. We consider three different groups of attack strategies:
\begin{enumerate}
    \item \emph{Greedy}~\cite{PaudiceML18} and our new attack \emph{Timber} always iterate over all the training instances for $k$ rounds and pick every time the one leading to the highest accuracy loss upon label flipping at each round. Timber exploits sub-tree retraining to improve efficiency. Timber is guaranteed to produce the same accuracy loss as the Greedy attack strategy, but it is expected to be faster in practice.  
    \item \emph{Greedy with early stopping (GES)} and our \emph{Timber with early stopping (TES)} iterate over all the training instances for $k$ rounds and, at each round, terminate as soon as they encounter an instance leading to some accuracy loss upon label flipping. TES exploits sub-tree retraining and processes instances in increasing order of score, prioritizing those where sub-tree retraining is more efficient.
    \item \emph{Entropy}~\cite{ZhangCZL21} and \emph{K-Medoids}~\cite{ZhangCZL21} are model-agnostic, black-box poisoning attacks operating in a single round. Entropy chooses the $k$ instances to flip according to a score based on the entropy measure, while K-Medoids separates the training instances into two clusters and chooses the $k$ instances to flip based on a mathematical distance.
\end{enumerate}

For all the strategies in the groups 1 and 2, we assume that if none of the processed instances introduces some accuracy loss, then the attacker flips the one leading to the smallest increase in accuracy. Although this choice goes against the attacker's goal in the short term, it may lead to model changes, enabling new attacks. Moreover, this choice forces all the attack strategies to always flip $k$ instances, leading to a fair comparison. \revise{We also experimented with a simple baseline based on random label flips~\cite{AnisettiABBDY23, YerlikayaB22, CoreyNB20}, but we chose to omit it due to its poor performance for the budget we consider, especially in comparison to our proposed attacks.}

We measure the effectiveness of different attack strategies in a white-box setting. We first use grid search to find the best model trained on the clean data. After the attack, we train a new model over the poisoned dataset using the same hyperparameters to estimate the accuracy loss. This setting represents a pessimistic scenario where the attacker has perfect knowledge of the training data and hyperparameters of the target model. We compare the effectiveness of the poisoning attacks in terms of F1 score and accuracy loss. We also consider the F1 score since the distribution of the class labels of the considered datasets is unbalanced. Even though Timber optimizes the loss of accuracy of the model, it also affects the F1 score by inducing misclassifications.

We focus on attacking decision tree ensembles, in particular RFs (without bootstrap sampling), since they are normally employed for tabular data classification in place of individual decision trees. The attacker tunes the number of trees from 2 to 15 and the maximum depth of the trees from 2 to 25. More details on the best RF model for each dataset are provided in the appendix.

    
    

\subsection{Attack Efficiency}
\label{sec: attack-efficiency}
We first assess the efficiency of the considered attacks by measuring the running time required to poison $10\%$ of the training set. This serves as an upper bound for the time needed to poison smaller subsets. In our experiments, we set a timeout of ten hours for each attack. The experiments have been performed on a virtual machine with 98 GB of RAM and Ubuntu 20.04.6 LTS, running on a server with an Intel Xeon Gold 6348 2.60GHz. To reduce running times, we rely on parallel implementations of the different attack strategies, using 16 threads. Training instances to poison are allocated to the different threads in a round-robin fashion. 

Table~\ref{tab: ensembles-runtimes} presents the running times of Greedy, Timber, GES and TES across different datasets. The black-box attacks K-Medoids and Entropy are much faster and always complete within a few seconds, so their runtimes are not comparable to white-box strategies and are not reported in the table to improve readability. Despite their efficiency, K-Medoids and Entropy are significantly less effective than the other attack methods, as reported in Section~\ref{sec: effectiveness-f1}. Our experiments confirm that Timber and TES are faster than their counterparts Greedy and GES. Remarkably, the Greedy attack strategy turned out to be infeasible on the Musk2 dataset, exceeding our timeout of ten hours. The computed speedup on the Musk2 and Spambase datasets ranges from 2x to 6x, while it is smaller (less than 2x) on the Wine dataset. This demonstrates the significant advantage in efficiency enabled by tree annotation and sub-tree retraining, that enable performing an attack over the entire training set in a reasonable amount of time. We finally observe that Timber and TES require around the same time as Greedy and GES to complete the attack on Breast-Cancer.


\begin{table}[t]
\centering
\caption{Runtime of the poisoning attacks with budget $k$ equal to $10\%$ of the training set size. Bold represents the best results in the two groups of columns.}
\begin{tabular}{c||c|c||c|c}
\toprule
\textbf{Dataset} & \multicolumn{4}{c}{\textbf{Runtime}} \\ 
\cmidrule(lr){2-5}
 & \textbf{Greedy} & \textbf{Timber} & \textbf{GES} & \textbf{TES} \\
\midrule
Musk2 & $>$10h & \textbf{1h41m} & 3h8m & \textbf{29m17s} \\
Wine & 3h21m & \textbf{2h59m} & 1h24m & \textbf{1h15m} \\
Spambase & 6h55m & \textbf{3h7m} & 1h12m  & \textbf{14m39s}\\
Breast & \textbf{2m38s} & 3m23s & 55s & \textbf{37s}  \\
\bottomrule
\end{tabular}
\label{tab: ensembles-runtimes}
\end{table}

    
    

\begin{figure*}[t]
  \centering
    \subfloat[]{\includegraphics[width=.245\textwidth]{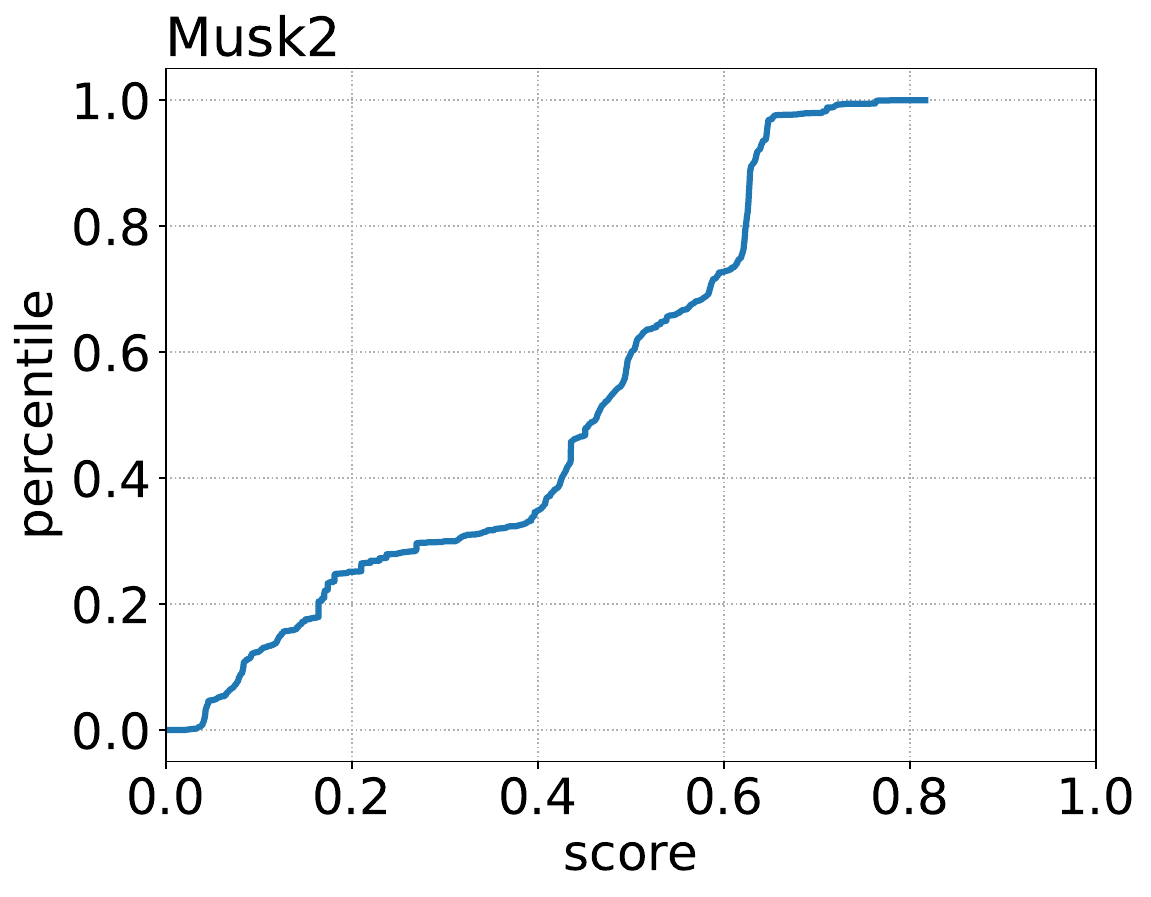}}
  \hfill
    \subfloat[]{\includegraphics[width=.245\textwidth]{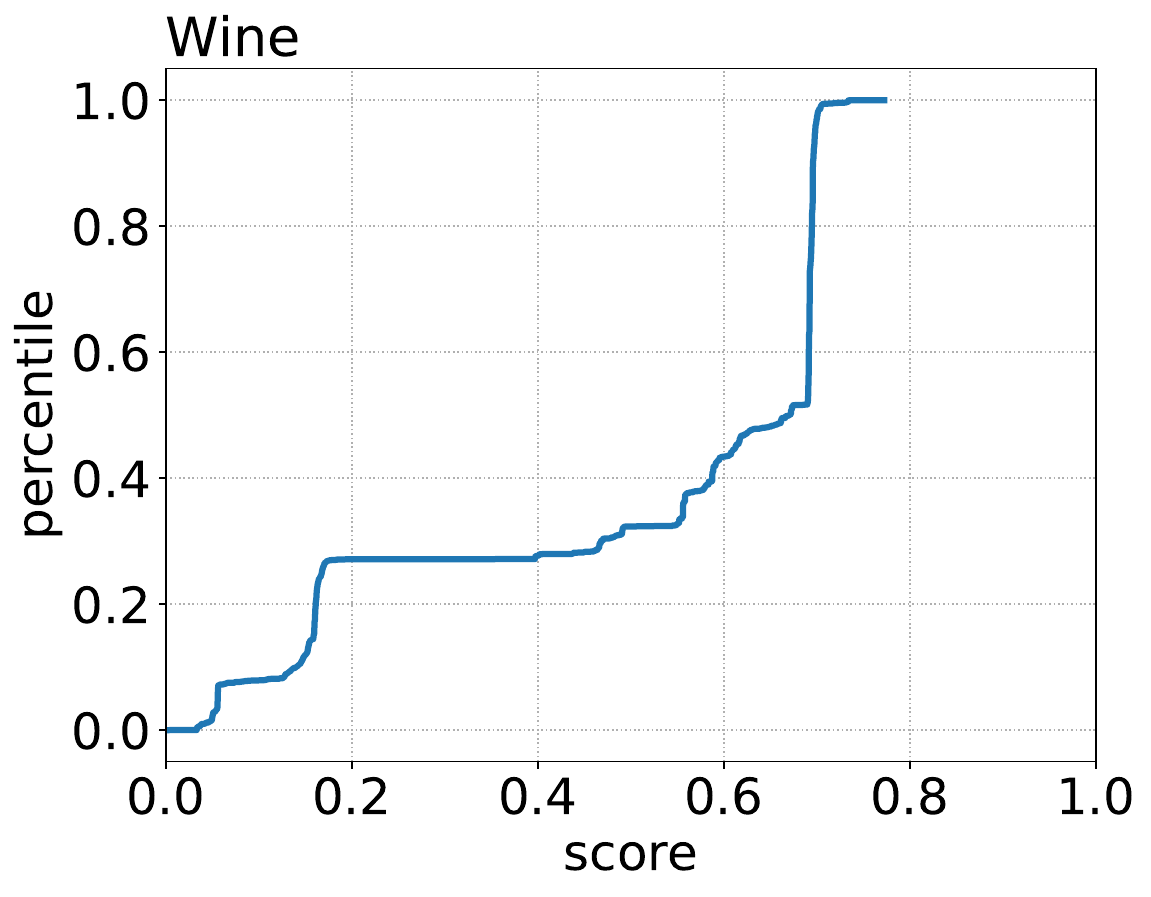}}
    \hfill
    \subfloat[]{\includegraphics[width=.245\textwidth]{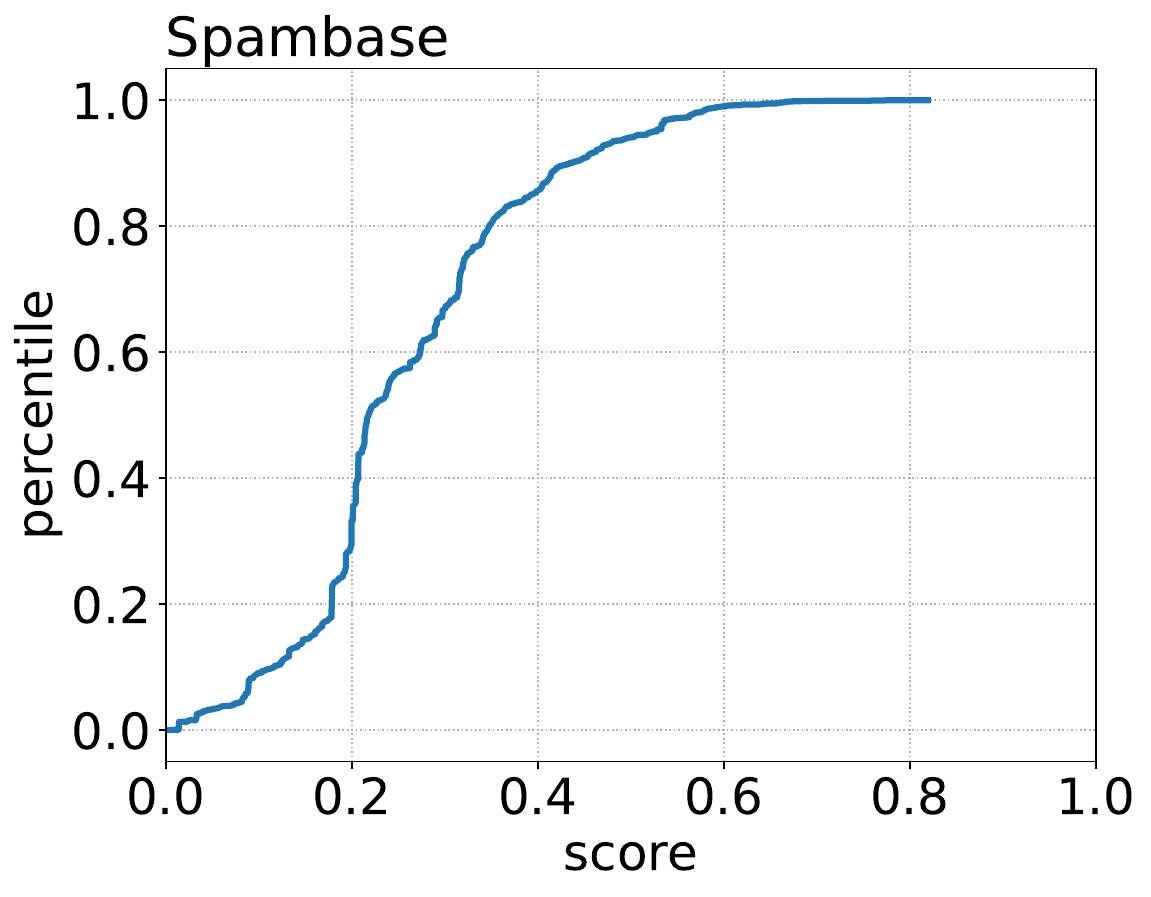}}
    \hfill
    \subfloat[]{\includegraphics[width=.245\textwidth]{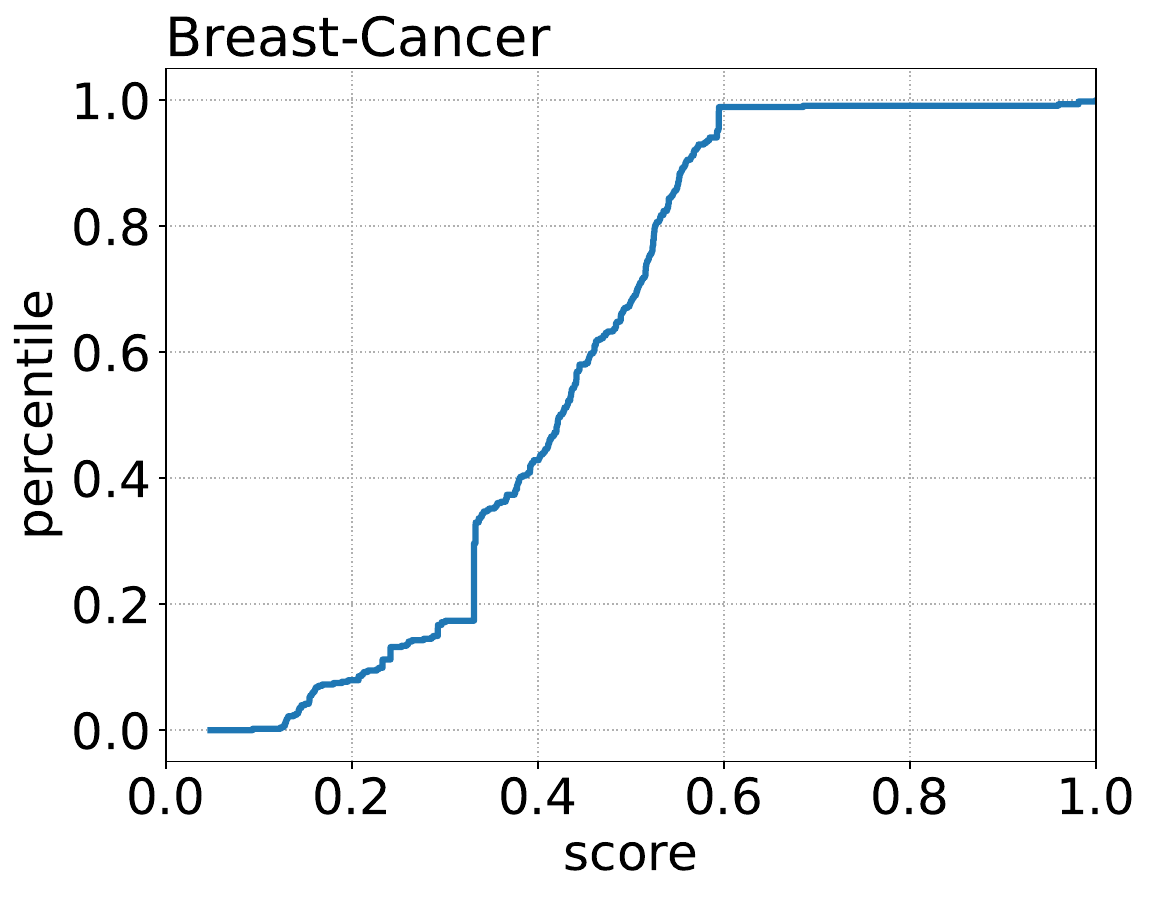}}
  \caption{Empirical cumulative distribution function of the mean scores of the training instances averaged over the rounds of TES on the considered datasets. The scores range from 0 to 1.}
  \label{fig: w-distr-tes}
\end{figure*}

We investigate the reasons behind the different speed-ups achieved by Timber and TES over Greedy and GES. We here focus on TES and we refer to the appendix for a similar discussion on Timber. Figure~\ref{fig: w-distr-tes} shows the empirical cumulative distribution function of the mean scores of each instance of the training set, averaged over the $k$ rounds of the TES attack.
On Musk2 and Spambase, more than $60\%$ of the mean scores are below 0.5, meaning that most of the training instances that can be attacked are located at the root of sub-trees with few training instances on average, i.e., less than $50\%$ of the number of instances in the training set, leading to high efficiency gains. This is particularly evident on the Musk2 dataset, the considered dataset with more instances and features, where TES is six times faster than GES. In contrast, the fact that less than $30\%$ of the instances have a mean score smaller than 0.5 on Wine motivates the higher runtime of TES on the dataset, where the overhead induced by the annotate and sub-tree retraining is less effectively compensated. Finally, although the majority of the mean scores in Breast-Cancer fall between 0.3 and 0.6, the dataset is too small to observe a considerable speedup. This is reasonable, since Timber and TES are designed to enable greedy poisoning attacks on large datasets. When the dataset is small and Greedy terminates in a few minutes, the optimizations introduced by our attacks may be unneeded. 

\subsection{Attack Effectiveness}
\label{sec: effectiveness-f1}

\begin{figure*}[t]
  \centering
    \subfloat[]{\includegraphics[width=.362\textwidth]{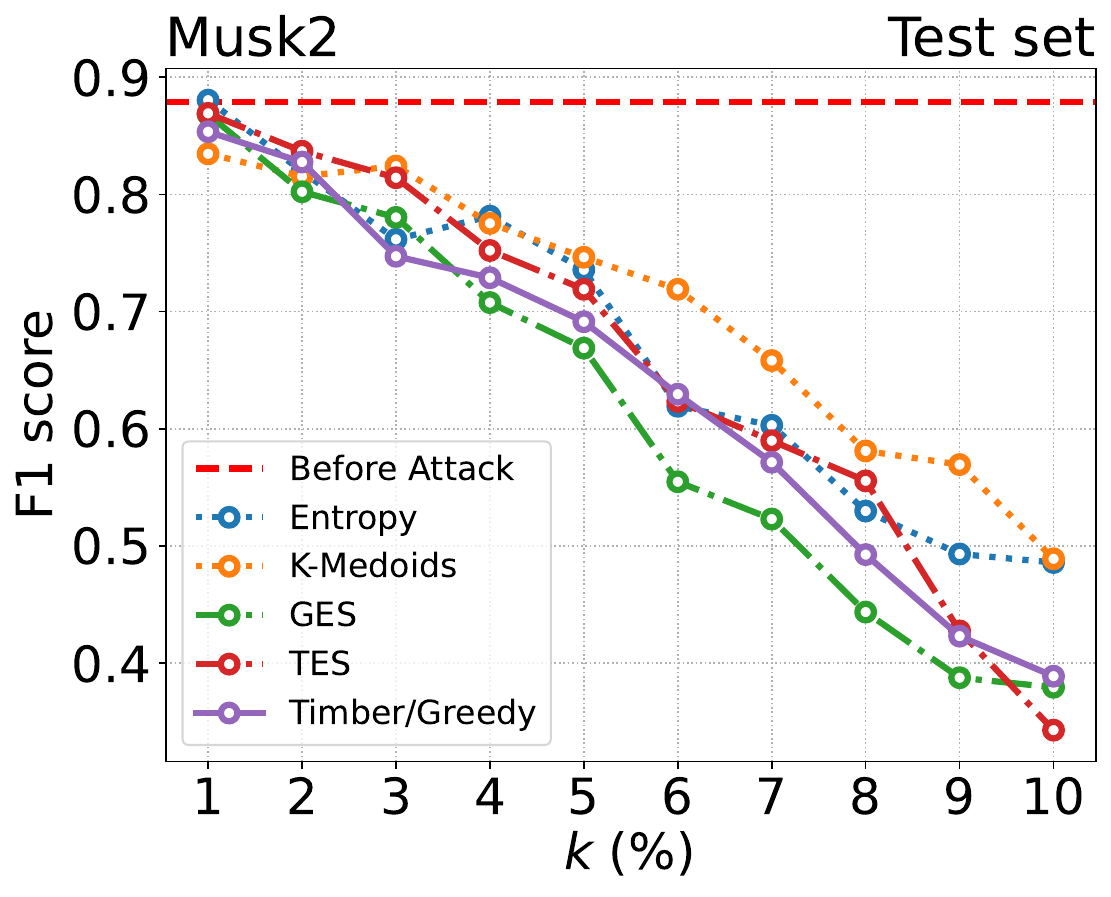}}
   \subfloat[]{\includegraphics[width=.362\textwidth]{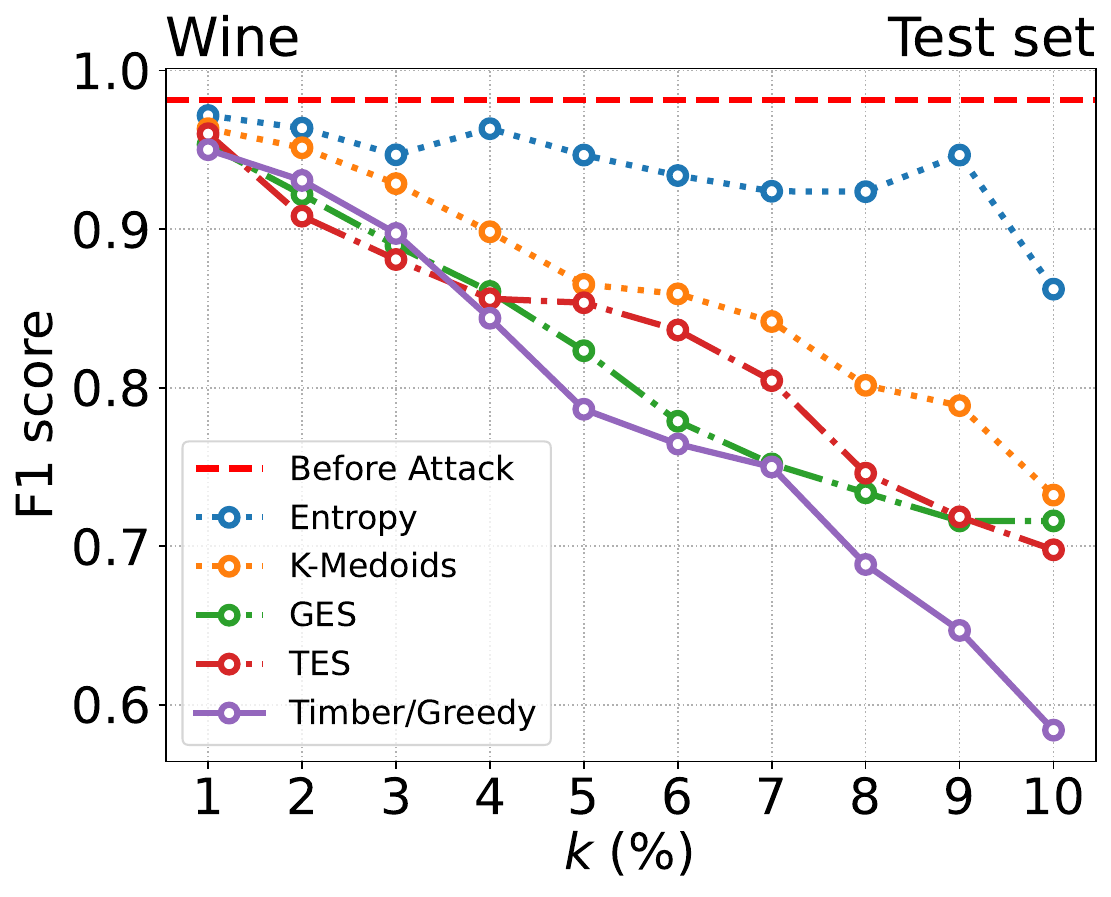}}
    \hfill
    \subfloat[]{\includegraphics[width=.362\textwidth]{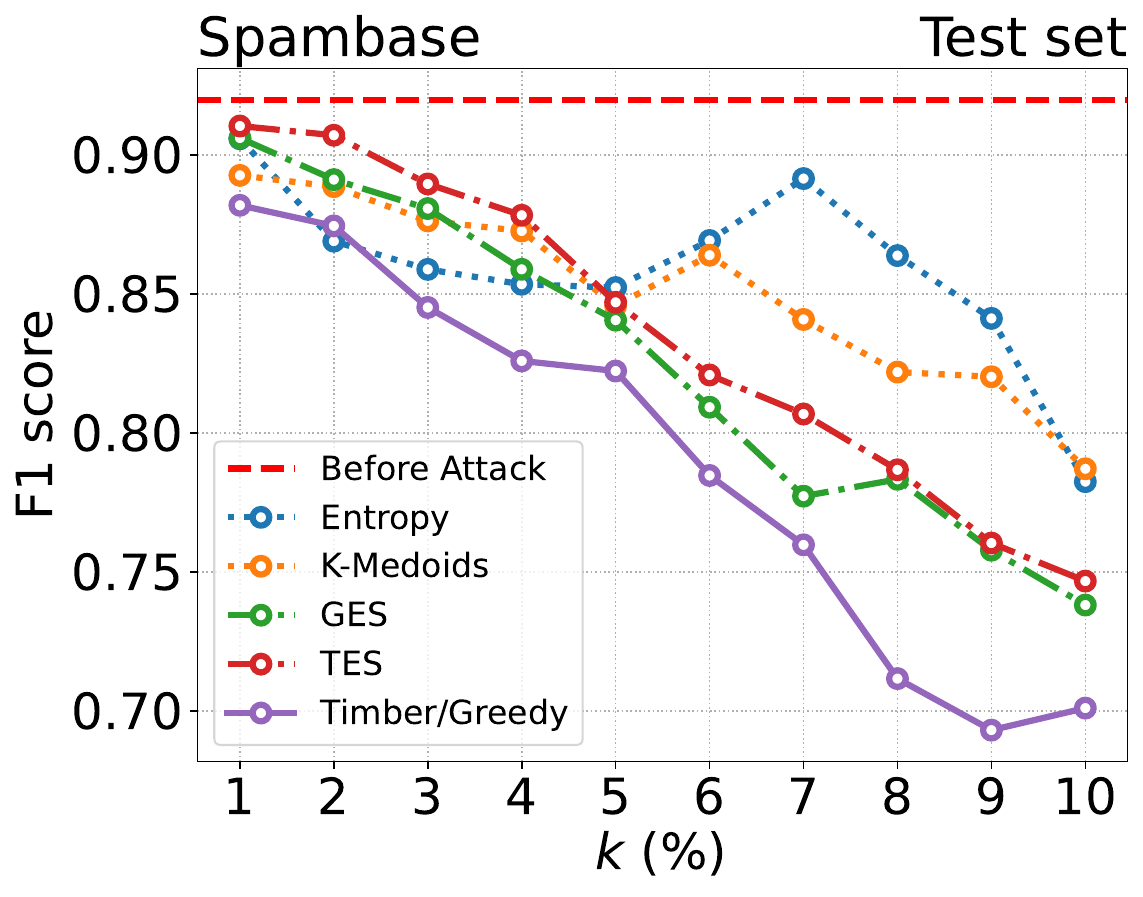}}
    \subfloat[]{\includegraphics[width=.362\textwidth]{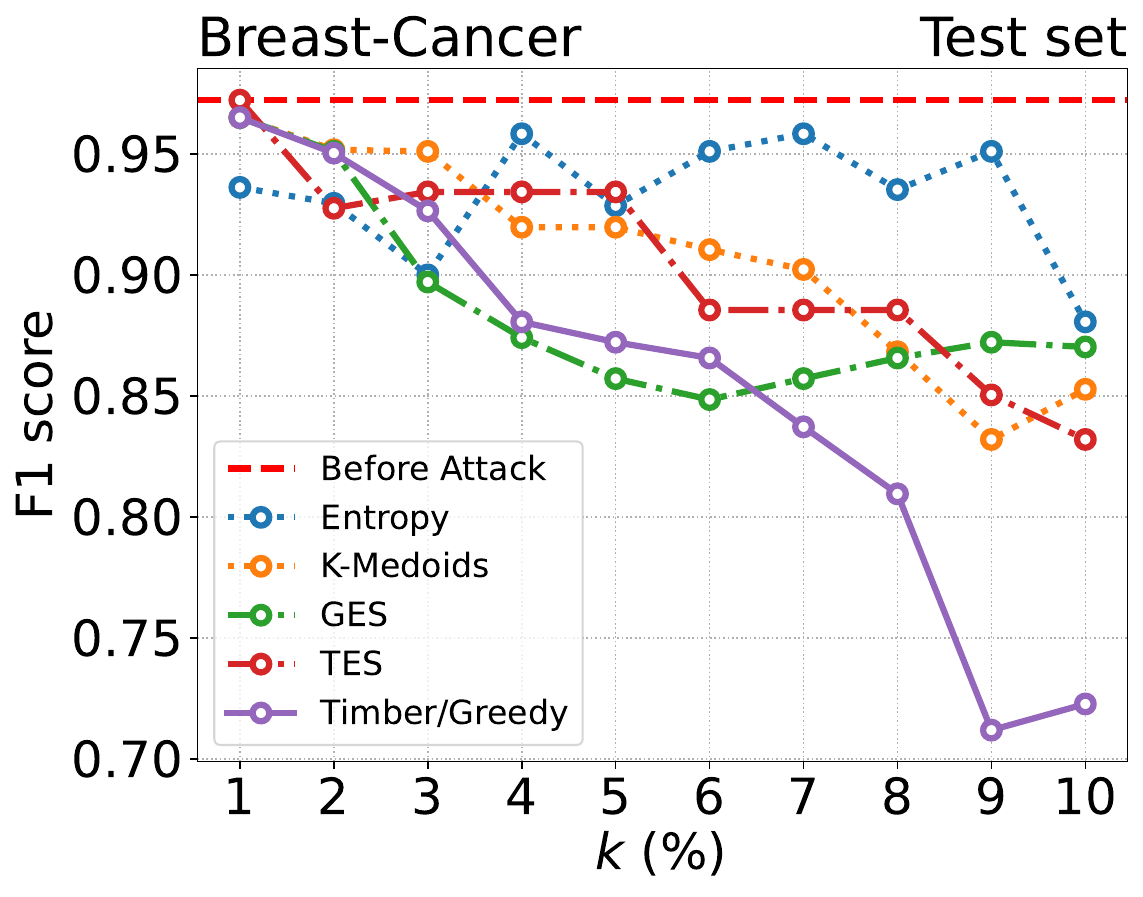}}
  \caption{F1 score of the attacked model under different poisoning attacks for budget $k$ equal to different percentages of poisoned training data, from 1\% to 10\%. A red horizontal line represent the F1 score of the model trained on the clean training set. Note that Timber is guaranteed to produce the F1 score loss as the Greedy attack strategy.}
  \label{fig: loss-f1-tree}
\end{figure*}

We now assess the effectiveness of the poisoning attacks in terms of F1 score loss on the test set. Figure~\ref{fig: loss-f1-tree} shows the F1 score loss induced by the considered attacks for each dataset and different values of $k$ (we just report a single line for Timber and Greedy, because they always produce the same output). We observe that the black-box attacks Entropy and K-Medoids are consistently outperformed by the other attacks. For example, on the Musk2 dataset, the initial F1 score is 0.88 and TES reduces it to 0.34, while the most effective black-box attack Entropy reduces it just to 0.49 (+0.15 over TES). Additionally, Timber/Greedy, which always iterates over all the training instances, performs better than the early-stopping attacks TES and GES on Wine, Spambase, and Breast-Cancer. This is expected since early-stopping attacks explore only a subset of attack options during each round. For instance, Timber/Greedy reduces the F1 score on Wine to 0.58, while TES is less effective, reducing it to 0.70 (+0.12). It may occasionally happen that attacks relying on early stopping are more effective in reducing the F1 score than Timber/Greedy, because all the considered attacks are greedy. This happens for TES on the Musk2 dataset, which reduces the F1 score to 0.34, while Timber reduces the F1 score of the model to 0.39 (+0.05). Finally, note that TES is more effective than GES on three datasets out of four, and it is only one point less effective than GES on Spambase. Its enhanced efficacy is likely due to sorting the instances exploited by the attack to improve efficiency. Attacking instances with lower scores, i.e., retraining sub-trees in which few training instances fall, allows the attack to perform more local changes, inducing consistent losses in the performance of the target model. Ultimately, Timber and TES are the most effective attacks, with Timber generally being more effective than TES in reducing the F1 score but at a higher computational cost. 

We can observe the same trends in the effectiveness of the poisoning attacks when considering the accuracy loss on the test set instead. For space reasons, we report the accuracy loss for each dataset and different values of $k$ in the appendix.

\subsection{Defenses}
\label{sec: defenses}
\revise{Our previous evaluation showed that Timber and TES attacks perform best, as Timber is usually the most effective and TES is usually the most efficient. We now show that both Timber and TES are effective attacks even when applying a defense against poisoning before training. We recall that poisoning attacks against decision trees have been under-explored and the same applies to defenses. We then focus on two model-agnostic defenses:} 
\begin{enumerate}
    \item \revise{\textit{kNN-based defense}~\cite{PaudiceML18}, a popular defense based on k-Nearest Neighbours (kNN) that performs \textit{training data sanitization}. For each instance of the possibly poisoned training set, it computes its $N$ nearest neighbors. If the fraction of the neighbors having the same label is greater than a threshold $\eta$, the algorithm assigns to the instance the label of these neighbors.} 
    The algorithm can be iterated for $M$ times on the training set.
    \item \revise{\textit{Bagging-based defense}~\cite{ChenLLYW22}, a recent defense based on bagging that performs \textit{robust training}.} It uses the defended classifier as a base classifier. It employs a variant of bagging based on hash functions to generate $G$ subsets of the possibly poisoned training set, each containing $K$\% instances of the training set. Then, it trains $G$ base classifiers on these subsets. The prediction for a test instances is obtained by aggregating the predictions of the base classifiers using hard majority voting.
\end{enumerate} 

In our evaluation, we perform grid search over the hyperparameters to select the values providing the highest F1 score on the validation set. This way, we estimate the effectiveness of the defense in the best possible setting from the defender's perspective. We consider $N \in \{4, 8, 12\}$, $\eta \in \{0.6, 0.75, 0.9\}$, $M \in \{1,3,5\}$, $G \in \{5, 10, \dots, 35, 40\}$ and $K \in \{20\%, 10\%, 5\%, 2.5\%$\}, including also the values used in the work presenting the defenses~\cite{PaudiceML18, ChenLLYW22}.

\begin{table}[t]
\centering
\caption{F1 score on the test set of the best model trained on different training sets: $F_1^c$ of the model trained on the clean training set, $F_1^p$ of the model trained on the poisoned training set, $F_1^n$ of the model trained on the training set sanitized by the kNN-based defense and $F_1^b$ of the model trained by the bagging-based defense on the poisoned training set.}
\begin{tabular}{c||c||c|c|c||c|c|c}
\toprule
\multirow{2}{*}{\textbf{Dataset}} & \multirow{2}{*}{$F_1^c$} & \multicolumn{3}{c}{\textbf{Timber}} & \multicolumn{3}{c}{\textbf{TES}}\\
\cmidrule{3-8}
& & $F_1^p$ & $F_1^{n}$ & $F_1^b$ & $F_1^p$ & $F_1^{n}$ & $F_1^b$ \\
\midrule
Musk2         & 0.88 & 0.39 & 0.55 & 0.50 & 0.34 & 0.56 & 0.48\\
Wine          & 0.98 & 0.58 & 0.81 & 0.80 & 0.70 & 0.83 & 0.84 \\
Spambase      & 0.92 & 0.70 & 0.78 & 0.86 & 0.75 & 0.78 & 0.81 \\
Breast & 0.97 & 0.72 & 0.94 & 0.97 & 0.83 & 0.93 & 0.95\\
\bottomrule
\end{tabular}
\label{tab:defence-stats-f1}
\end{table}

To understand the effectiveness of the evaluated defenses, we compute four measures over the test set: the F1 score of the original model trained on the clean training set (denoted by $F_1^c$), the F1 score of the model trained over the poisoned dataset (denoted by $F_1^p$) and the F1 score of the model trained on the poisoned dataset after applying the defense (denoted by $F_1^d$, with $d \in \{n,b\}$ discriminating between the kNN-based defense and the bagging-based defense). This allows us to compute for each defense $d$ the \emph{estimated defense benefit} $F_1^d - F_1^p$, i.e., the increase in F1 score enabled by the application of the defense w.r.t. the undefended poisoned model, and the \emph{estimated residual damage} $F_1^c - F_1^d$, i.e., the decrease in F1 score w.r.t. the original model 

The computed results are reported in Table~\ref{tab:defence-stats-f1}. The numbers show that the analyzed defenses provide some mitigation against our attacks. In particular, the estimated defense benefit $F_1^n - F_1^p$ for the kNN-based defense ranges between 0.03 and 0.23, with an average value of 0.15, while the estimated defense benefit $F_1^b - F_1^p$ for the bagging-based defense ranges between 0.06 and 0.25, with an average value of 0.15. Nevertheless, the damage caused by our poisoning attacks despite the application of the defenses is significant. The estimated residual damage $F_1^c - F_1^n$ for the kNN-based defense ranges between 0.03 and 0.33, with an average value of 0.17, while the estimated residual damage $F_1^c - F_1^b$ for the bagging-based defense ranges between 0 and 0.40, with an average of 0.16. This implies that, on average, our attack reduces the $F_1$ score of the original model by between 0.16 and 0.17, even when one of the two defenses is applied. Thus, the analyzed defenses effectively mitigate the damage of our poisoning attacks, but they are far from being able to completely thwart them. The only dataset where the application of the defense yields a model with comparable performance to the original one is Breast-Cancer. This can be explained by the simplicity of this dataset, where a decision stump (i.e., a decision tree of depth 1) achieves an $F_1$ score of 0.95. This suggests that the classes are easily distinguishable, enhancing the effectiveness of the defenses. \revise{These observations are confirmed by looking at the effect of the defenses on the accuracy, that we report in the appendix.}

    


\section{Related Work}
We here discuss poisoning attacks and defenses using the taxonomy provided in~\cite{CinaGDVZMOBPR23}.

\subsection{Poisoning Attacks}


\textit{Availability} poisoning attacks aim to degrade the accuracy of the target classifier to compromise its utility. Existing label flip attacks and our new attacks, Timber and TES, belong to this category, which assumes that the attacker can only modify the labels of the instances in the training set. The objective is to find the combination of flips that leads to the best accuracy loss of the target model. Label flip attacks have been deeply investigated for support vector machines~\cite{BiggioCFGR11, XiaoXE12, XiaoBNXER15, PaudiceML18}, linear regression models~\cite{AwasthiBL17} and neural networks~\cite{ZhangBHRV17, ZhangHSW20}. To the best of our knowledge, no work in the literature has proposed poisoning attacks specifically for decision tree ensembles. Previous work~\cite{BarrenoNJT10, AnisettiABBDY23, CoreyNB20, ShahidIWIA22, AryalGA22, YerlikayaB22} evaluates the robustness of decision tree ensembles against the random label flip attack, a model-agnostic attack that selects the label to flip randomly. \cite{ZhangCZL21} proposes other two model-agnostic attacks, \textit{Entropy} and \textit{K-Medoids}, that are also evaluated on decision tree ensembles.
Our work fills an important gap in the literature by proposing the first poisoning attack specifically tailored for decision tree ensembles, which is feasible on large datasets and clearly outperforms other attack approaches.

Other availability poisoning attacks are clean-label, i.e., they assume that the attacker can modify only features~\cite{BiggioNL12, XiaoBBFER15, Munoz-GonzalezB17, TaltySB21, PrudHommeK21}, and hybrid, in the sense that they target both features and labels~\cite{FengCZ19, abs-2103-02683, MeiZ15}. Bilevel poisoning attacks are the most popular attacks of these two categories. They find the best perturbation to apply to the training data by solving a bilevel optimization problem~\cite{BiggioNL12}. Most of the attacks of this type
target differentiable models since they exploit gradients
extracted from the loss function of the target model
to find the best perturbations to apply to the training
instances. However, decision trees are non-differentiable models. Thus, previous work evaluated only model-agnostic clean-label and hybrid attacks~\cite{VerdeM021, TaltySB21, PrudHommeK21, AnisettiABBDY23} on decision tree ensembles. Designing clean-label and hybrid poisoning attacks for decision tree ensembles is a relevant direction for future work.

Finally, another category of poisoning attacks aims at harming the \textit{integrity} of ML models~\cite{Munoz-GonzalezB17, KohL17, ShafahiHNSSDG18, GuoL20, HuangGFTG20, GuLDG19, LiuMALZW018, NguyenT21, SarkarBKGM22}. The objective is to preserve the general performance of the target model, while causing the misclassification of specific samples. 
The aim of these attacks is different from ours, since our two proposed attacks target the availability of decision tree ensembles, so we do not compare against this category. However, even integrity attacks against decision tree ensembles have not been deeply investigated in the literature. Designing efficient and effective integrity-poisoning attacks against decision tree ensembles is another interesting line of research for future work.

\subsection{Defenses Against Poisoning}
We focus on defenses against availability poisoning attacks, in particular label flip attacks, grouped into two classes: \textit{training data sanitization} and \textit{robust training}~\cite{CinaGDVZMOBPR23}.

Training data sanitization defenses are model-agnostic approaches that remove poisoning samples from the training set before training by recognizing instances that are different from the other legitimate training points. Previously proposed techniques exploit k-Nearest Neighbours classifiers~\cite{PaudiceML18}, clustering algorithms~\cite{LaishramP16} and outlier detection algorithms~\cite{FredericksonMDP18, SteinhardtKL17}.

Robust training defenses aim instead at mitigating the effect of poisoning during training. In particular, the defenses consist of training algorithms that mitigate the effect of poisoned samples. Some model-agnostic defenses of this type exploit bagging, leveraging the observation that using small subsets of the training set for training ensembles can mitigate the effect of poisoning~\cite{BiggioCFGR11, JiaCG21, 0001F21, 00020F22, ChenLLYW22, AnisettiABBDY23}. These defenses and the Randomized Smoothing-based defense~\cite{RosenfeldWRK20} can also provide certificates about the robustness of the model to availability poisoning attacks. Another defense proposed in~\cite{Nelson2009} removes instances from the training set if they induce a significant loss in accuracy when used in training. Finally, specific defenses for differentiable models have been proposed and exploit robust optimization~\cite{DiakonikolasKK019, LiuLVO17, JagielskiOBLNL18}, regularization~\cite{BiggioNL11, DemontisBFGR17} and loss correction~\cite{PatriniRMNQ17}. 

To the best of our knowledge, no robust training defenses have been specifically designed to protect decision tree ensembles, even though all the model-agnostic defenses previously described can be applied. Algorithms that verify the robustness of decision trees against poisoning attacks have been proposed instead~\cite{DrewsAD20, MeyerAD21}.~\cite{AnisettiABBDY23} is the only work specifically evaluating the application of a defense to decision tree ensembles, in particular RFs, against availability poisoning attacks like random label flip. It employs the hash bagging defense inspired by previous work~\cite{0001F21, 00020F22, ChenLLYW22}, and it observes a degradation of the performance of the RFs even when adopting the defense. However, it considers unrealistic attack budgets ranging from 10\% to 30\% of training instances. Our work is orthogonal to this work since we do not focus on defenses, but we propose new attacks specifically tailored for decision tree ensembles. We demonstrate that our attacks Timber and TES are still effective when one representative defense~\cite{PaudiceML18, ChenLLYW22} from each group is adopted, even when we consider more realistic attacker's capabilities, i.e., the attacker can flip at most 10\% of the labels of the training set. The two defenses partially mitigate the effect of the attacks, but they are not able to thwart them. 
We leave the evaluation of other defenses as future work, as well as designing defenses specifically tailored to enhance the robustness of decision tree ensembles to poisoning attacks.





\section{Conclusion}

We presented Timber, the first white-box poisoning attack for decision trees. 
Timber uses a greedy strategy, incorporating an annotation procedure for the tree and sub-tree retraining to efficiently assess the impact of poisoned instances, optimizing the computational cost of the attack. This allows Timber to scale to larger datasets than standard greedy methods. The Timber variant with early stopping offers faster runtimes, though with potentially reduced effectiveness. We also extended Timber to decision tree ensembles, particularly random forests, to demonstrate its relevance in real-world machine learning applications. Our experiments on public datasets show that Timber and its variant with early stopping outperform existing black-box strategies in terms of attack effectiveness and existing greedy attacks in terms of attack efficiency. Moreover, our two attacks are not thwarted by two representative defenses.

\revise{As future work, we would like to generalize our techniques to GBDT models and design more powerful defenses for poisoning attacks specific to decision tree ensembles. We also plan to study how to efficiently perform clean-label and integrity poisoning attacks against decision tree ensembles, to fill the gap in the literature.}

\paragraph*{Acknowledgements}
This research was supported by project SERICS (PE00000014) under the MUR National Recovery and Resilience Plan funded by the European Union - NextGenerationEU. Moreover, it acknowledges support from the European Union - Next-GenerationEU - PNRR – M.4 C.2, I.1.1 - PRIN 2022 WHAM!, 2022ZZX57L, H53D23003750006.

\bibliographystyle{IEEEtran}
\bibliography{biblio}

\newpage
\appendix

\revise{In this section, we present figures and tables that include information and experimental results omitted from the main body of the paper.}

\revise{We first show in Table~\ref{tab: datasets} the key characteristics of the datasets and in Table~\ref{tab: ensembles-stats} the details of the best RF model for each dataset.}

\begin{figure*}[t]
  \centering
    \subfloat[]{\includegraphics[width=.245\textwidth]{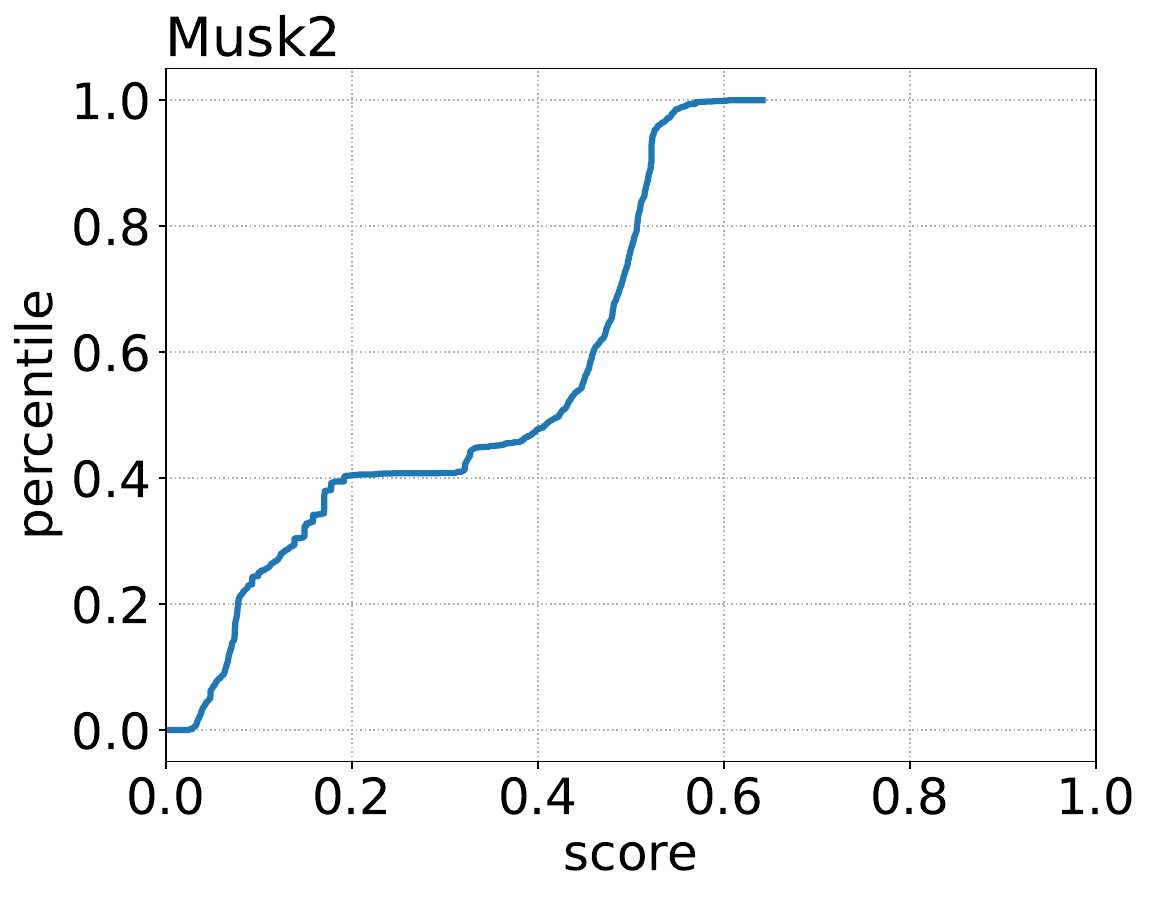}}
  \hfill
    \subfloat[]{\includegraphics[width=.245\textwidth]{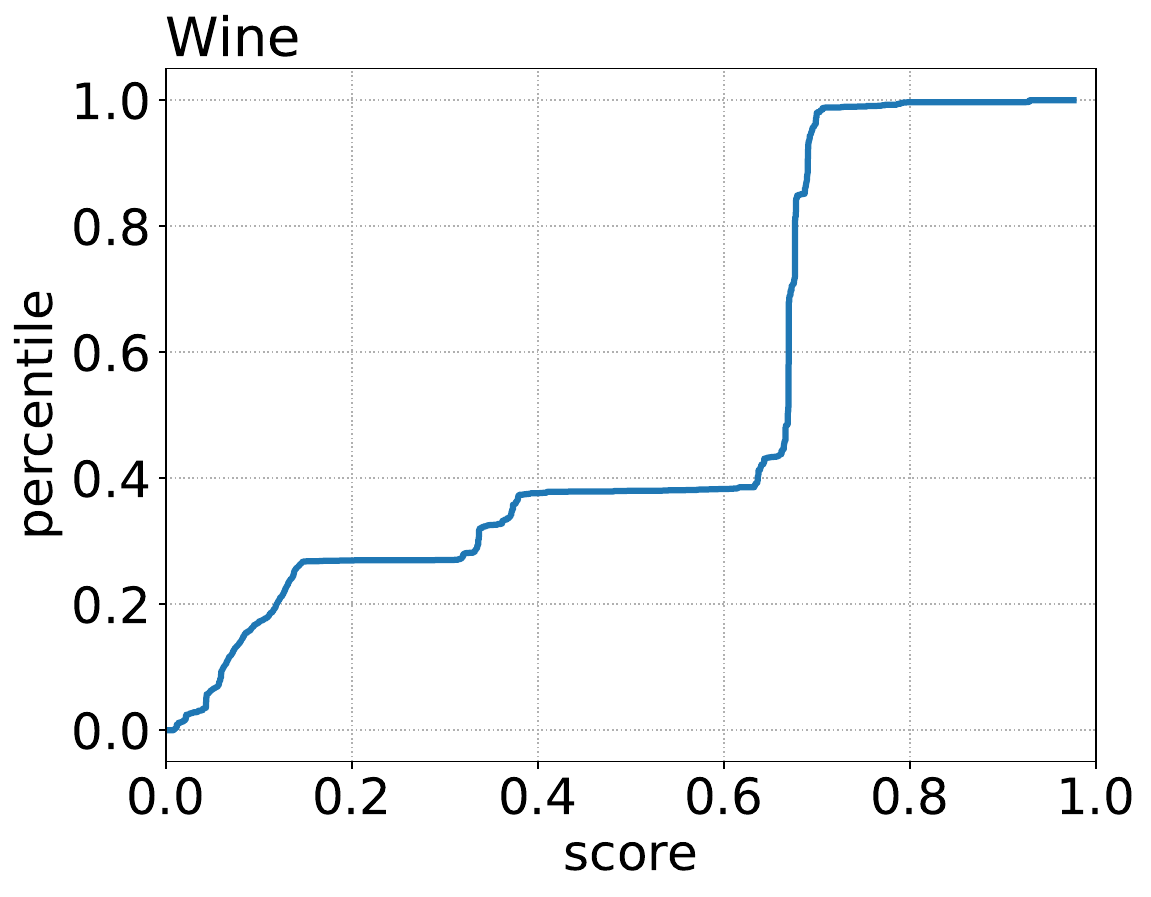}}
    \hfill
    \subfloat[]{\includegraphics[width=.245\textwidth]{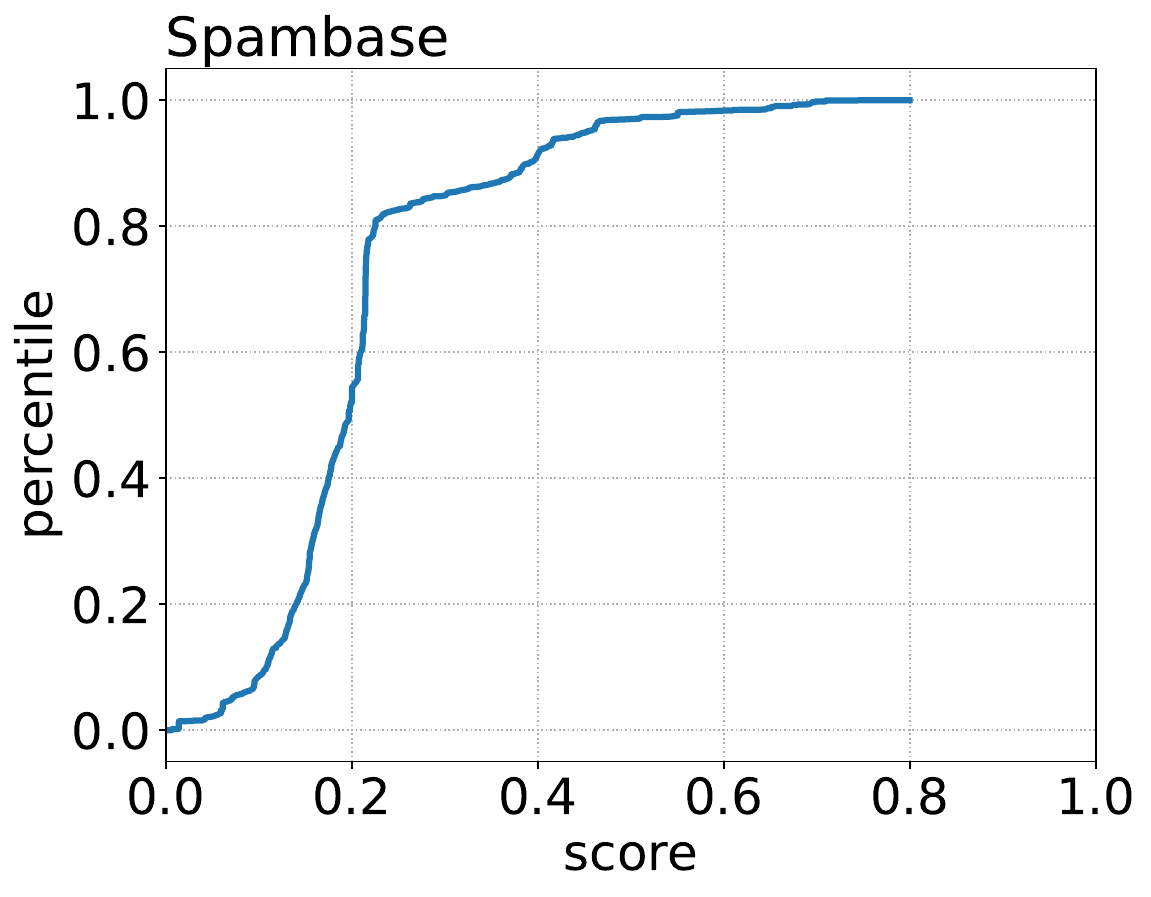}}
    \hfill
    \subfloat[]{\includegraphics[width=.245\textwidth]{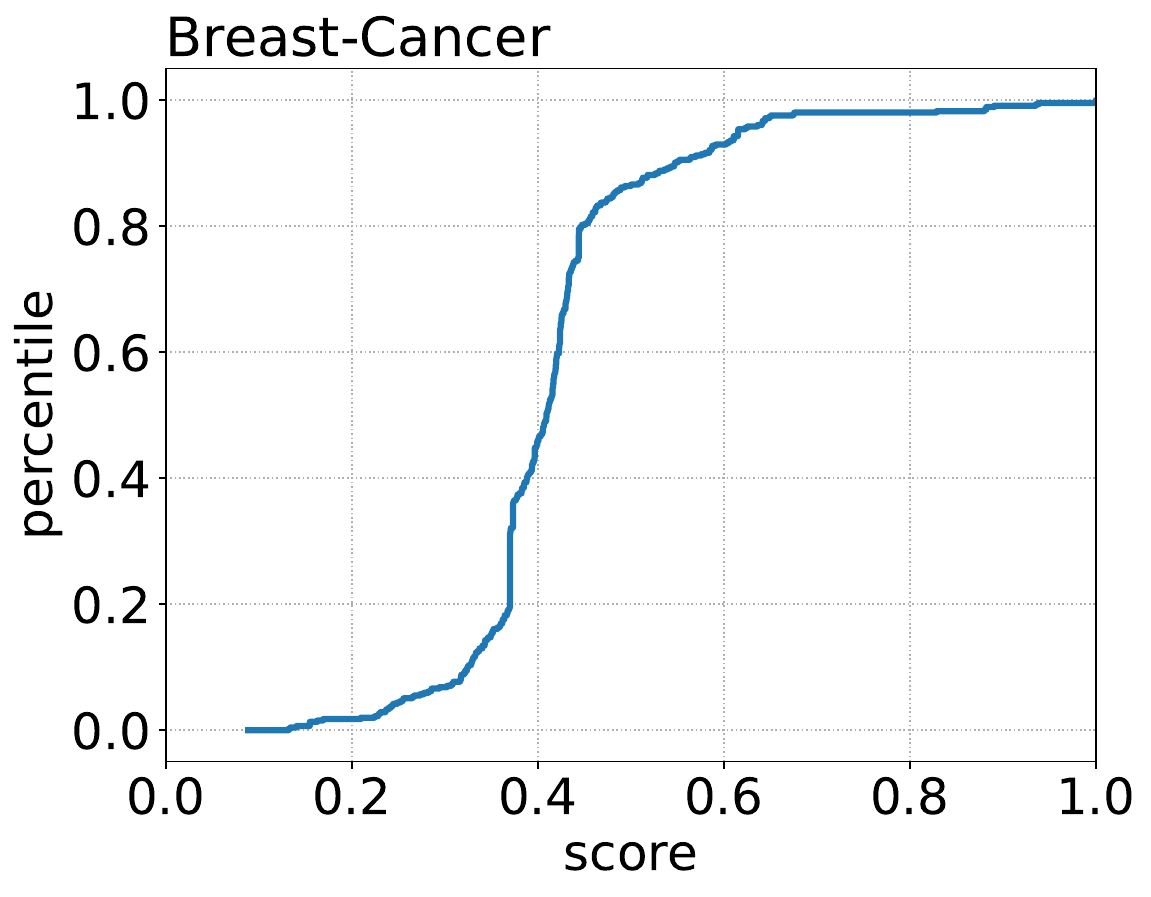}}
  \caption{Empirical cumulative distribution function of the mean scores of the training instances over the iterations of Timber on the considered datasets. The scores range from 0 to 1.}
  \label{fig: w-distr-timber}
\end{figure*}

We then show in Figure~\ref{fig: w-distr-timber} the empirical cumulative distribution function of the mean scores of each instance in the training set, averaged over $k$ rounds of the Timber attack, where $k$ is set to 10\% of the training set size. The distributions of the mean scores across the datasets are similar to those observed for the TES attack. Therefore, the reasoning behind the speed-up of Timber over Greedy follows the same rationale as the speed-up of TES over GES \revise{(see Section~\ref{sec: attack-efficiency})}.

\begin{figure*}[t]
  \centering
    \subfloat[]{\includegraphics[width=.36\textwidth]{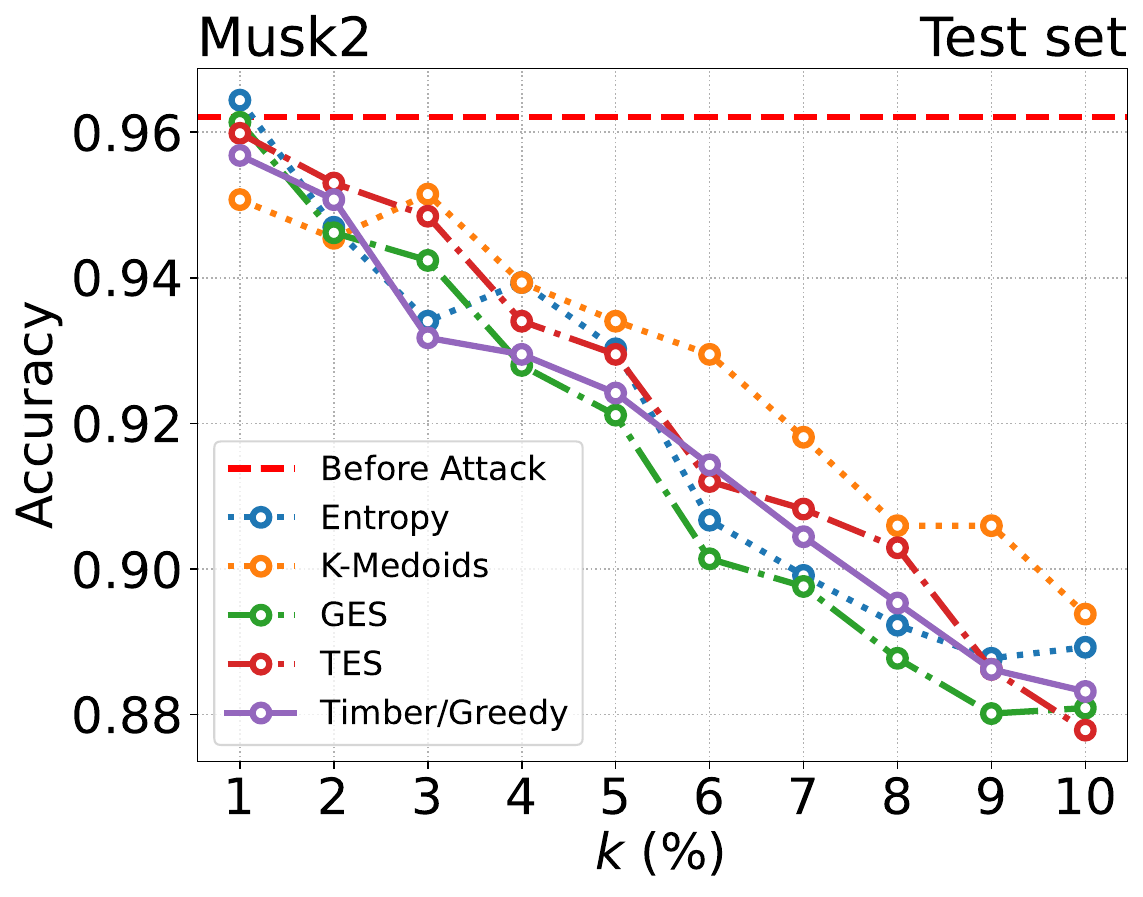}}
 \subfloat[]{\includegraphics[width=.36\textwidth]{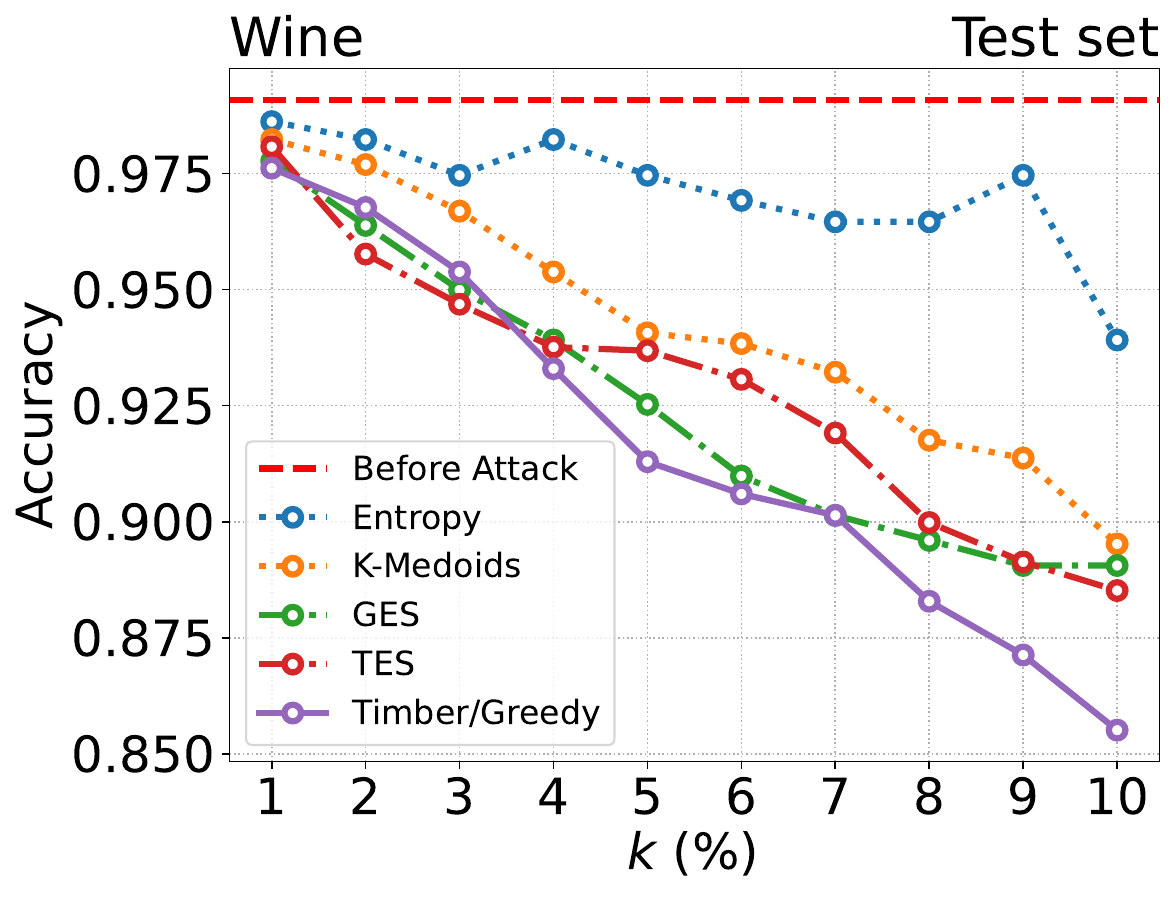}}
  \hfill
    \subfloat[]{\includegraphics[width=.36\textwidth]{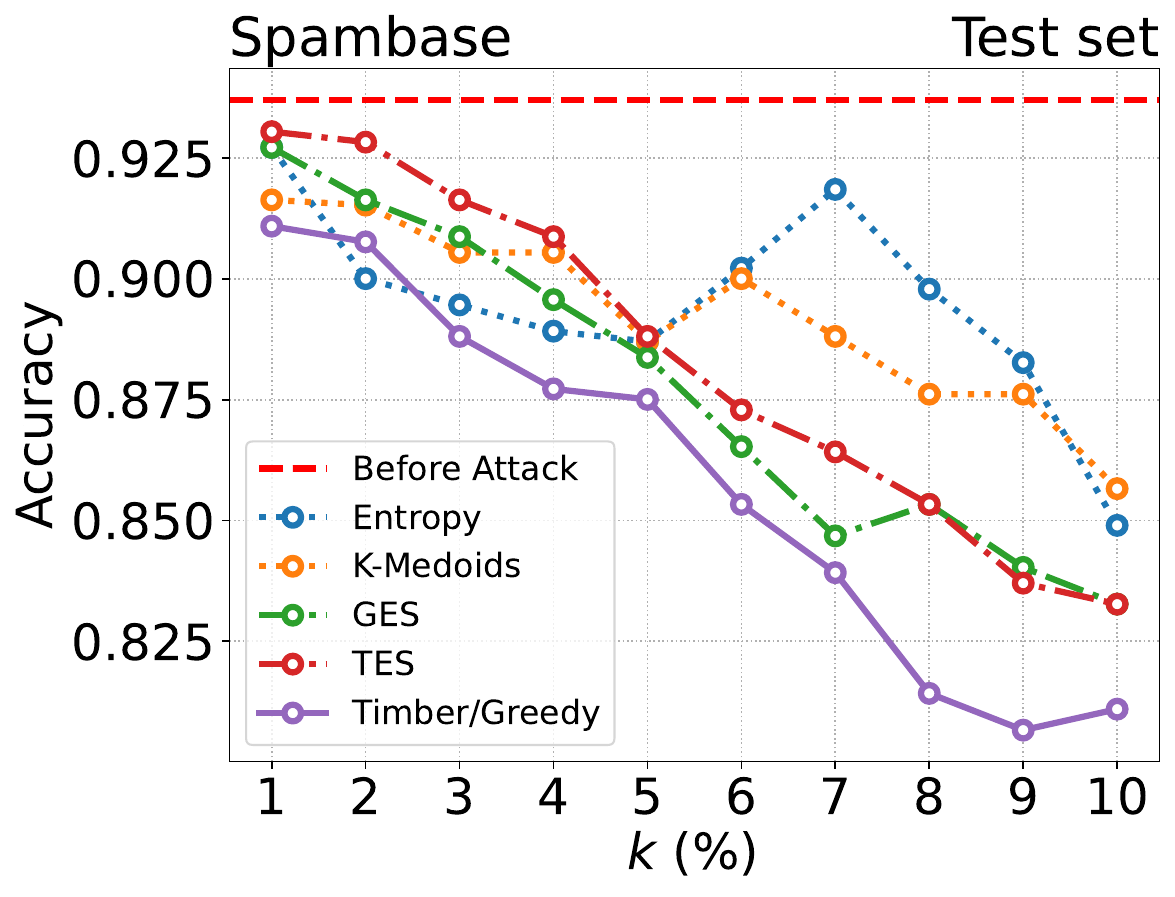}}
    \subfloat[]{\includegraphics[width=.36\textwidth]{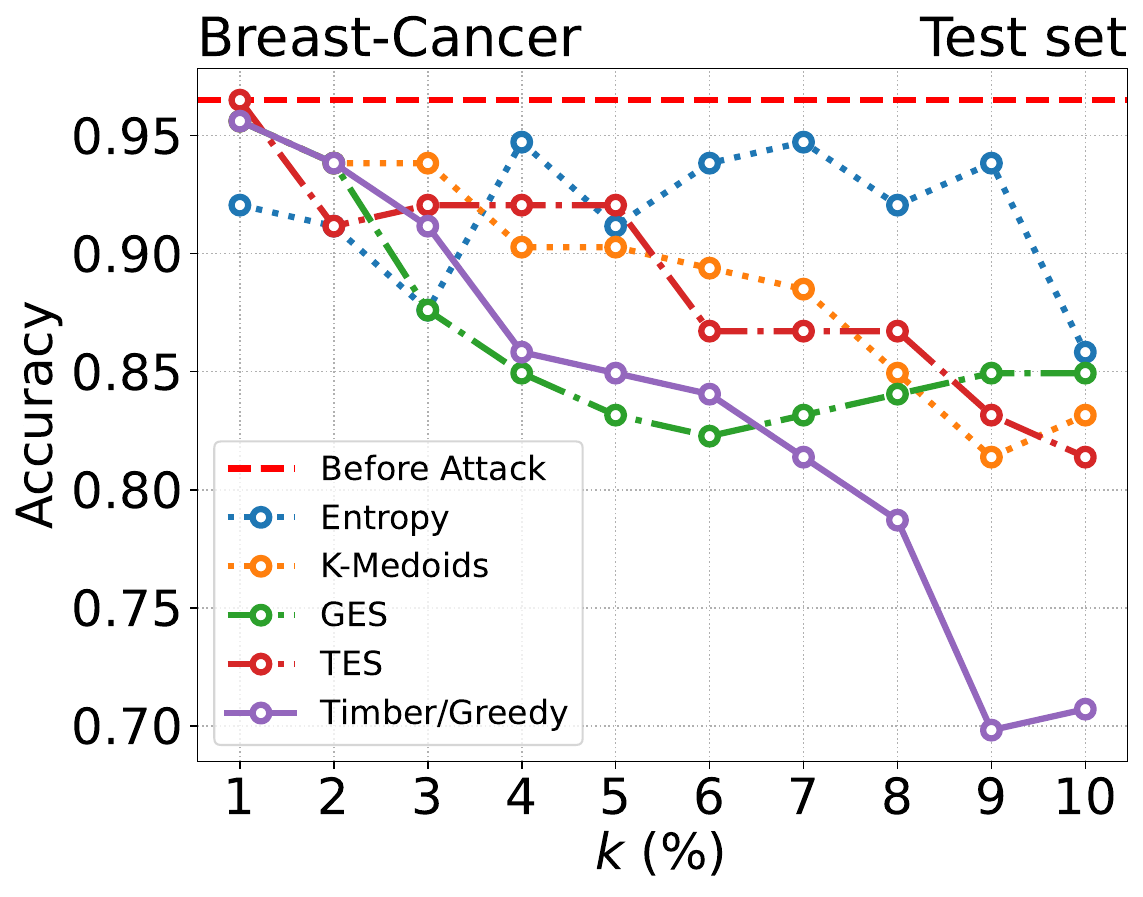}}
 \caption{Accuracy of the attacked model under different poisoning attacks for budget $k$ equal to different percentages of poisoned training data, from 1\% to 10\%. A red horizontal line represent the accuracy of the model trained on the clean training set. Note that Timber is guaranteed to produce the same accuracy loss as the Greedy attack strategy.}
  \label{fig: loss-acc-test}
\end{figure*}

\revise{Figure~\ref{fig: loss-acc-test} instead shows the accuracy loss on the best model for each dataset, induced by the pool of considered attacks across different values of $k$.} The trends in the results are similar to those observed for the F1 score loss, confirming the insights into the effectiveness of the attacks derived from the discussion on the F1 score loss induced by each attack \revise{(see Section~\ref{sec: effectiveness-f1})}.

Finally, we show in Table~\ref{tab: defence-stats-acc} the effectiveness of the evaluated defenses, using accuracy as the performance metric. In particular, we compute four measures over the test set: the accuracy of the original model trained on the clean training set (denoted by $a^c$), the accuracy of the model trained over the poisoned dataset created by Timber or TES (denoted by $a^p$) and the accuracy of the model trained on the poisoned dataset after applying the defense (denoted by $a^d$, with $d \in \{n,b\}$ discriminating between the kNN-based defense and the bagging-based defense). 
The results align with those derived from the F1 score evaluations \revise{(see Section~\ref{sec: defenses})}, showing that the defenses can mitigate the impact of both attacks but cannot actually thwart them.

\vspace{3cm}

\begin{table}[ht]
    \centering
    \caption{Dataset statistics.}
    \begin{tabular}{c|c|c|c}
    \toprule
    \textbf{Dataset} & \textbf{Instances} & \textbf{Features} & \textbf{Distribution} \\
    \midrule
    Musk2 & 6,598 & 166 & $85\% / 15\%$ \\
    Wine & 6,497 & 11 & $75\% / 25\%$ \\
    Spambase & 4,601 & 57 & $61\% / 39\%$ \\
    Breast & 569 & 30 & $63\% / 37\%$\\
    \bottomrule
    \end{tabular}
\label{tab: datasets}
\end{table}

\begin{table}[ht]
\centering
\caption{Number of trees, maximum depth, accuracy and F1 score on the test set of the RF.}
\begin{tabular}{c|c|c|c|c}
\toprule
\textbf{Dataset} & \textbf{\# Trees} & \textbf{Max. Depth} & \textbf{Accuracy} & \textbf{F1 score}\\
\midrule
Musk2 & 7 & 20 & 0.96 & 0.88 \\ 
Wine & 14 & 9 & 0.99 & 0.98 \\
Spambase & 15 & 20 & 0.94 & 0.92\\
Breast & 14 & 7 & 0.97 & 0.97  \\
\bottomrule
\end{tabular}
\label{tab: ensembles-stats}
\end{table}

\begin{table}[ht]
\centering
\caption{Accuracy on the test set of the best model trained on different training sets: $a^c$ of the model trained on the clean training set, $a^p$ of the model trained on the poisoned training set, $a^n$ of the model trained on the training set sanitized by the kNN-based defense and $a^b$ of the model trained by the bagging-based defense on the poisoned training set.}
\begin{tabular}{c||c||c|c|c||c|c|c}
\toprule
\multirow{2}{*}{\textbf{Dataset}} & \multirow{2}{*}{$a^c$} & \multicolumn{3}{c}{\textbf{Timber}} & \multicolumn{3}{c}{\textbf{TES}}\\
\cmidrule{3-8}
& & $a^p$ & $a^{n}$ & $a^b$ & $a^p$ & $a^{n}$ & $a^b$ \\
\midrule
Musk2         & 0.96 & 0.88 & 0.90 & 0.89 & 0.88 & 0.90 & 0.88 \\ 
Wine          & 0.99 & 0.86 & 0.92 & 0.91 & 0.89 & 0.93 & 0.93 \\
Spambase      & 0.94 & 0.81 & 0.85 & 0.90 & 0.83 & 0.85 & 0.87\\
Breast & 0.96 & 0.71 & 0.92 & 0.96 & 0.81 & 0.91 & 0.96 \\

\bottomrule
\end{tabular}
\label{tab: defence-stats-acc}
\end{table}

\end{document}